\documentclass{article}

\usepackage{microtype}
\usepackage{multicol}
\usepackage{multirow}
\usepackage{graphicx}
\usepackage{subcaption}
\usepackage{algorithmic}
\usepackage{booktabs} 
\usepackage{colortbl}
\usepackage{hyperref}

\usepackage[preprint]{icml2026}

\usepackage{amsmath}
\usepackage{amssymb}
\usepackage{mathtools}
\usepackage{amsthm}

\usepackage[capitalize,noabbrev]{cleveref}

\theoremstyle{plain}

\theoremstyle{definition}

\theoremstyle{remark}

\usepackage[textsize=tiny]{todonotes}

\begin{document}

\twocolumn[
  \icmltitle{DBGL: Decay-aware Bipartite Graph Learning for Irregular Medical Time Series Classification}




  \begin{icmlauthorlist}
    \icmlauthor{Jian Chen}{1}
    \icmlauthor{Yuzhu Hu}{2}
    \icmlauthor{Xiaoyan Yuan}{2}
    \icmlauthor{Yuxuan Hu}{3}
    \icmlauthor{Jinfeng Xu}{1}
    \icmlauthor{Yipeng Du}{1}
    \icmlauthor{Wenhao Yuan}{1}
    \icmlauthor{Wei Wang}{4}
    \icmlauthor{Edith C. H. Ngai*}{1}
  \end{icmlauthorlist}
  
  \icmlaffiliation{1}{Department of Electrical and Electronic Engineering, The University of Hong Kong, Hong Kong, China}
  \icmlaffiliation{2}{Shenzhen MSU-BIT University, China}
 \icmlaffiliation{3}{Department of Data Science, City University of Hong Kong, China}
  \icmlaffiliation{4}{Engineering Research Centre of Applied Technology on Machine Translation and Artificial Intelligence, Macao Polytechnic University}
 
  \icmlcorrespondingauthor{Jian Chen}{ccccccj03@connect.hku.hk}

  \icmlkeywords{Machine Learning, ICML}

  \vskip 0.3in
]



\printAffiliationsAndNotice{* corresponding author.}  

\begin{abstract}
  Irregular Medical Time Series play a critical role in the clinical domain to better understand the patient's condition. However, inherent irregularity arising from heterogeneous sampling rates, asynchronous observations, and variable gaps poses key challenges for reliable modeling. Existing methods often distort \textbf{temporal sampling irregularity} and missingness patterns, while failing to capture \textbf{variable decay irregularity}, resulting in suboptimal representations. To address these limitations, we introduce DBGL: Decay-Aware Bipartite Graph Learning for Irregular Medical Time Series. DBGL first introduces a patient–variable bipartite graph that simultaneously captures irregular sampling patterns without artificial alignment and adaptively models variable relationships for temporal sampling irregularity modeling, enhancing representation learning. To model variable decay irregularity, DBGL designs a novel node-specific temporal decay encoding mechanism that captures each variable's decay rates based on sampling interval, yielding a more accurate and faithful representation of irregular temporal dynamics. We evaluate the performance of DBGL on four publicly available datasets and results show that DBGL outperforms all baselines, and our code is also available in the supplementary material.
\end{abstract}

\section{Introduction}

Medical time series (MTS) are central to healthcare analysis but are often irregular due to clinical workflows, costs, and patient conditions \cite{sun2020review,chenderi}. This irregularity arises from heterogeneous sampling frequencies (e.g., continuous heart rate vs. hourly blood tests), asynchronous observations, and variable temporal intervals, resulting in irregular medical time series (IMTS) \cite{shukla2018modeling}. While it poses challenges for conventional modeling \cite{wu2022long}, it also contains informative signals: clinicians alter measurement frequency and ordering in response to concern or changing severity, and the resulting patterns of sampling and missingness (e.g., bursts around events, long gaps during stability) can be exploited as prognostic and diagnostic cues. Successfully addressing these challenges is of great significance for early risk prediction, patient state tracking, and so on \cite{tan2020explainable}.

Modeling IMTS is challenging not only because observations occur at heterogeneous rates, but also because these irregularities distort temporal dependencies. This temporal sampling irregularity \cite{nielsen1994analysis,liu2022itimes}, i.e., uneven and asynchronous measurements, complicates the alignment of variables in a coherent representation space. In addition, different clinical variables evolve at different rates over time, a phenomenon we term variable decay irregularity \cite{lipton2016modeling,che2018recurrent}. Formally, we quantify this effect using a decay rate $\lambda$ estimated from empirical autocorrelations of each variable: higher $\lambda$ indicates fast-changing variables (e.g., heart rate, blood pressure), while lower $\lambda$ corresponds to slower, more stable variables (e.g., creatinine, hemoglobin). This provides a unified, quantitative measure of clinically meaningful differences in temporal dynamics. Representative decay rates estimated on the P12 dataset are shown in Table~\ref{tb:lamb} and more details can be found in the Appendix~\ref{vda}.

\begin{table}[htbp] 
\centering
\caption{Representative variable decay rates ($\lambda$) on the P12 dataset.}
\vspace{-2mm}
\label{tb:lamb}
\scalebox{0.8}{
\begin{tabular}{l c l c}
\hline
Variable & $\lambda$ & Variable & $\lambda$ \\
\hline
HR        & 1.73    & Temp       & 0.0265 \\
Creatinine & 0.1399 & Lactate    & 13.00  \\
ALP       & 14.71   & PaCO2      & 0.0225 \\
ALT       & 14.62   & AST        & 13.89  \\
\hline
\end{tabular}
}
\end{table}

Irregular sampling is a central challenge in clinical time-series. Traditional interpolation or resampling pipelines produce regularly spaced sequences but may obscure informative missingness patterns \cite{chen2018neural}. Recent RNN-based \cite{che2018recurrent,rajkomar2018scalable}, diffusion-based \cite{rubanova2019latent,kidger2020neural}, and transformer-based approaches \cite{zheng2024irregularity,ren2024periormer} incorporate continuous-time embeddings or learned interpolation to mitigate these issues, yet they still operate primarily on flattened sequences and therefore struggle to represent which variable is observed at which time, a key structural signal in clinical data. Bipartite variable–time graphs \cite{yalavarthi2024grafiti,liu2025timecheat} make this structure more explicit by linking variable nodes to time nodes, but they are typically constructed within local windows and do not capture patient-specific sampling behaviors. Moreover, clinical variables naturally exhibit heterogeneous temporal dynamics \cite{tan2020data}, some change within minutes, others over hours or days, yet most existing models encode time uniformly and lack explicit mechanisms for variable-specific decay.

In summary, existing approaches face two main limitations. First, preprocessing or flattened sequence modeling can obscure inherent irregularity: resampling may remove informative missingness, and sequence-based variable modeling captures limited asynchronous patient-variable interactions. Second, temporal dynamics are often treated uniformly, ignoring that clinical variables evolve at heterogeneous rates and exhibit variable-specific sensitivity to elapsed time. This can result in incomplete or less informative patient representations for downstream tasks.

To address these limitations, we propose \textbf{D}ecay-aware \textbf{B}ipartite \textbf{G}raph \textbf{L}earning (\textbf{DBGL}) with two key components. First, for temporal sampling irregularity, DBGL models IMTS as a sequence of patient–variable bipartite graphs, where at each time step, edges are created only for observed variables. This design preserves the true observation structure and explicitly encodes irregular sampling patterns without artificial alignment. Second, we introduce a node-specific temporal decay encoding mechanism, which decays patient hidden states according to the elapsed time and then updates them with new variable observations. This enables each variable to “forget” at clinically realistic speeds while maintaining a continuously evolving patient state. We evaluate DBGL on four publicly available clinical datasets. Results demonstrate that DBGL consistently outperforms all baselines across tasks. Detailed analyses and ablation studies further reveal the key factors driving its performance gains. Our main contributions can be summarized as:
\begin{itemize}
    \item We propose DBGL, a novel framework that embeds irregular sampling patterns directly into the graph topology. By constructing each time step as a patient–variable bipartite graph, DBGL preserves informative observation dependencies without artificial alignment. Through graph message passing, correlations among variables are adaptively aggregated into patient nodes, yielding more expressive and patient-specific representations.
    \item We introduce a novel temporal decay encoding mechanism with node-specific updates to model variable-dependent decay irregularities in IMTS. Unlike a uniform temporal discount, our design allows each variable to follow its own adaptive decay trajectory. At each time step, the hidden state decays according to the sampling interval and is then updated based on new observations. This design captures fine-grained and heterogeneous temporal dynamics across variables for a continuously evolving representation.
    \item We conduct comprehensive experiments on four public clinical datasets. Experimental results demonstrate that DBGL achieves superior performance compared to existing methods on all datasets, showcasing its robustness and adaptability across various scenarios.
\end{itemize}

\section{Related Work}
\subsection{Irregular Time Series Analysis}
Irregularly sampled time series are common in healthcare, finance, and transportation. Existing approaches mainly fall into two categories. Interpolation-based methods, such as kernel smoothing, Gaussian processes \cite{tan2021cooperative}, or temporal aggregation \cite{ma2020adacare}, resample irregular observations onto regular grids, but often distort original sampling patterns and obscure informative missingness. Direct modeling approaches avoid resampling by explicitly incorporating time intervals, including recurrent networks for non-uniform gaps \cite{che2018recurrent}, temporal embeddings for arbitrary timestamps \cite{horn2020set,shuklamulti}, and neural ODEs for continuous-time dynamics \cite{schirmer2022modeling,chen2023contiformer}. Attention- and graph-based architectures have also been explored for capturing long-range dependencies and inter-variable relations. Yet, most prior work focuses on local irregularities, leaving asynchronous variable dependencies and variable-specific decay largely unmodeled.

\subsection{Graph-based Methods for Clinical Time Series}
Graph neural networks (GNNs) naturally capture dependencies among variables, patients, and temporal contexts in clinical time series. Prior studies have applied GNNs to model inter-variable relations for prediction \cite{escudero2024explainable}, capture spatial-temporal EEG dependencies \cite{varatharajah2017eeg,tangself}, reflect dependencies among clinical variables \cite{zheng2023graph}, predict patient interventions \cite{xu2024predicting}, model dynamic multi-resolution temporal-spatial dependencies \cite{fan2025towards}, and integrate textual medical knowledge \cite{luo2024knowledge}. Despite these advances, such approaches do not explicitly address irregular medical time series, lacking mechanisms to leverage the inherent sampling and decay irregularities in IMTS. To bridge this gap, we introduce \textbf{DBGL}, a decay-aware bipartite graph learning framework that explicitly encodes temporal decay and sampling irregularities, enabling more robust and faithful representation learning for irregular clinical time series.

\section{Methodology}

\subsection{Overview}
\begin{figure*}
    \centering
\includegraphics[width=0.825\linewidth]{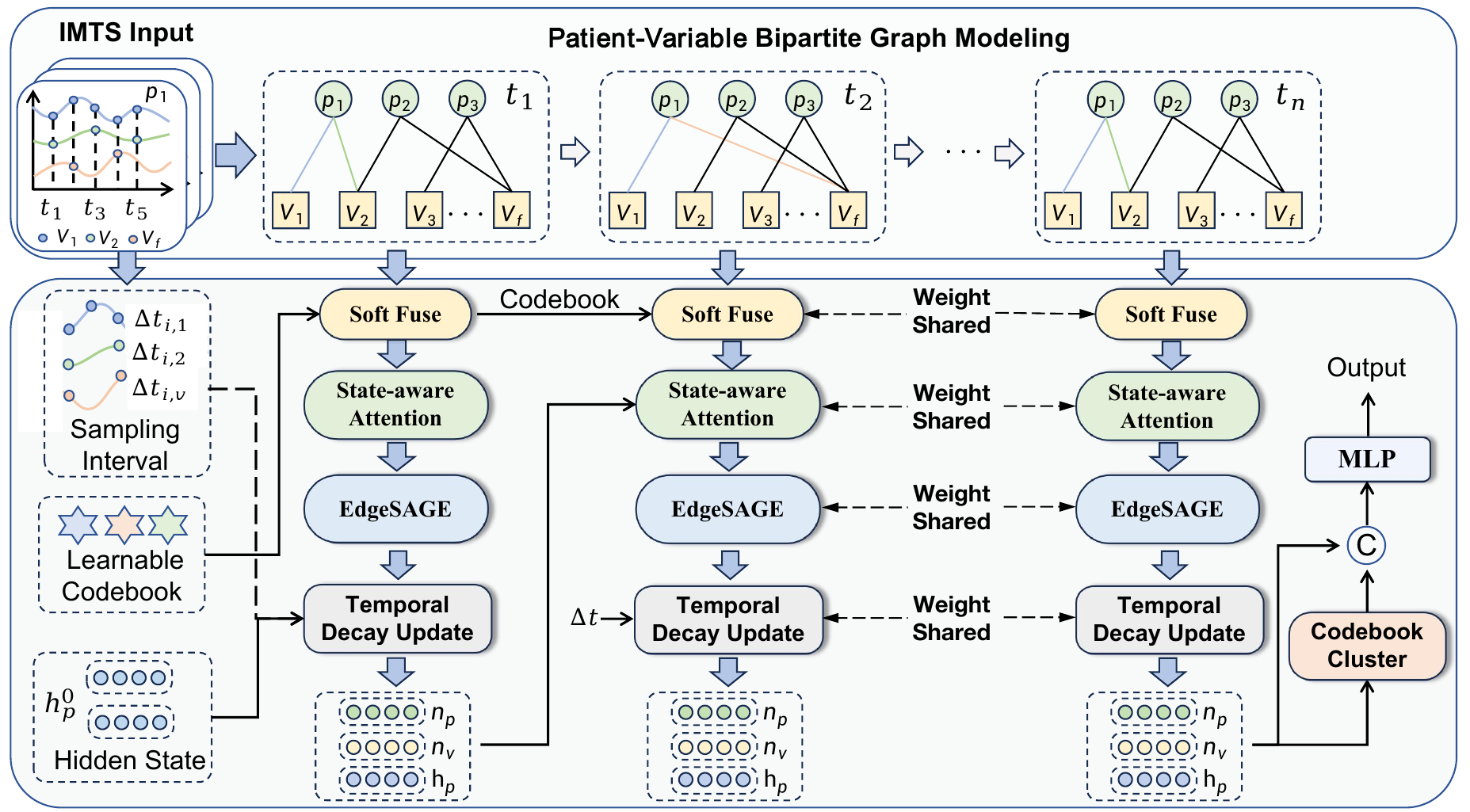}
    \vspace{-2mm}
    \caption{Framework of our proposed DBGL. DBGL first models the IMTS as patient-variable bipartite graphs. Furthermore, a novel mechanism called temporal decay encoding with node-specific updates is used to obtain a robust representation for classification.}
    \label{Framework}
\end{figure*}

DBGL addresses the irregular sampling of IMTS by modeling each time step as a patient–variable bipartite graph, where sampling patterns are naturally encoded in the graph structure without artificial temporal alignment. This design preserves informative inter-variable dependencies under non-uniform sampling. To capture variable-specific temporal dynamics, DBGL introduces a node-wise temporal decay encoding that adaptively modulates each variable’s influence over time. Furthermore, a learnable codebook is employed to model shared latent patient states, enabling more expressive and generalizable patient representations. The overall framework of DBGL is illustrated in Figure~\ref{Framework}.

\subsection{Patient-Variable Bipartite Graph}
Temporal sampling irregularity is a key challenge in MTS, as different variables are measured at heterogeneous and unpredictable intervals. Early methods typically resort to resampling or imputation, which may distort temporal dynamics and obscure informative structures. Recent graph-based methods attempt to address this issue by constructing fully connected variable graphs and masking adjacency entries according to observed variables. However, such masking mainly encodes observation availability and fails to effectively capture the sampling irregular patterns and model the interaction between patients and variables explicitly.

\textbf{Bipartite Patient-Variable Graph Construction.} To overcome these limitations, we represent medical time series as a sequence of patient-variable bipartite graphs. Specifically, for each time step, DBGL transforms the medical time series features $\{X^{p}\}^{N}_{p=1}$ and observation flag matrix $M$ to an undirected bipartite patient-variable
graph $G_{t} = (V, E)$. Here, the node set $V= V_{p}\ \cup V_{v}$, consisting of $p$ patient nodes $V_{p}=\{u_{1}, u_{2},...,u_{p}\}$ and $n$ variable nodes $V_{v} = \{v_{1},v_{2},...,v_{n}\}$. The edge set $E$ consists of the current observed state, which is determined using the observation flag matrix $M$, and is referred to as the adjacency matrix. Specifically, an edge $e_{u_p,v_n}$ will be constructed if the patient $p$ has the data for the variable $n$. In this way, the whole IMTS will be modelled as a series ographsph $G = \{G_{1}, G_{2},..., G_{T}\}$, where $T$ is the total time step.

\textbf{Embedding Initialization.} For each observation, we obtain an edge embedding by summing three components: a value embedding derived from the measurement, a time embedding encoding absolute sampling intervals via linear and sinusoidal projections \cite{kazemi2019time2vec}, and a learnable variable type embedding. Unobserved entries are masked out according to $M$. In addition, patient nodes are initialized as constant vectors during each forward pass, while variable nodes are learnable parameters. Together, these node and edge embeddings provide the bipartite graph with both observation-specific semantics and adaptable representations of patients and variables.

\textbf{Message Passing.}
Based on the constructed patient-variable bipartite graph, we employ a multi-layer EdgeSAGE network to propagate and aggregate information across patients, variables, and their temporal observations. At the $l$-th layer, messages are first constructed by jointly encoding neighbor node states and edge attributes as Eq.~\ref{eq:message}:
\vspace{-1mm}
\begin{equation}
m^{(l)}_{i,j} = \sigma \left(W_m^{(l)}[v_j^{(l)} ||e_{ij}^{(l)}] + b_m \right),
\widetilde{v}_{i}^{(l)} = \sum_{j\in N_{i}} m^{(l)}_{i,j},
\label{eq:message}
\end{equation}
\vspace{-0.5mm}
where $||$ denotes concatenation, $v_{j}^{(l)}$ is the embedding of neighbor $j$, $e_{ij}^{(l)}$ is the edge embedding of edge linking node $i$ and $j$, $\sigma$ is the ReLU activation, and $N_{i}$ is the neighbor set of node $i$. The final output of this module is the updated node and edge states, $v_{p,n}^{t}$ and $e_{p,n}^{t}$, as shown in Eq.~\ref{eq:update}:
\begin{equation}
\begin{aligned}
v^{(l+1)}_{i} 
&= \sigma\!\left(W_h^{(l)} [v_i^{(l)} \,\|\, m_{i,j}^{(l)}] + b\right), \\
e_{i,j}^{(l+1)} 
&= e_{i,j}^{(l)} + \sigma\!\left(W_h^{(l)} [v_i^{(l+1)} \,\|\, v_j^{(l+1)} \,\|\, e_{i,j}^{(l)}] + b\right).
\end{aligned}
\label{eq:update}
\end{equation}

Unlike conventional models that either resample or impute irregularly observed data, our message passing explicitly conditions edge embeddings on the actual observation times. This allows patient nodes to continuously integrate variable-specific dynamics at irregular intervals, while edge updates preserve temporal dependencies between observations. As a result, DBGL effectively mitigates the temporal sampling irregularity of IMTS by adaptively aligning patient trajectories with the true clinical observation patterns.

\subsection{Temporal Decay Encoding Mechanism}
\textbf{Temporal Decay Encoding.} To address the variable-specific temporal irregularity in IMTS, we design a temporal decay encoding mechanism that adaptively adjusts each variable’s hidden state according to its sampling interval. We first initialize the state representation of each variable for every patient, $h^{(0)}$, as zero, and then design a temporal decay encoding mechanism to update these states at each time step based on the representations aggregated from the bipartite graph and the previous hidden state. 

Specifically, at each time step $t$, given $h^{(t-1)}_{p,n}$, the previous hidden state of the $n$-th variable of the patient $p$, we compute a decay rate $\lambda_{p,n}^{t}$ based on the current edge representations $e_{p,n}^{t}$ through a small MLP, and introduce a sampling decay factor is then determined by the elapsed interval $\Delta t$ for hidden state decay factor $\gamma_{p,n}^{t}$, as illustrating in Eq.~\ref{eq:decay}:
\begin{equation}
\lambda_{p,n}^{t} = \text{Softplus}\left(\text{MLP}(e_{p,n}^{t})\right),\gamma_{p,n}^t = e^{-\lambda_{p,n}^{t} \cdot \Delta t},
\label{eq:decay}
\end{equation}
and then the decayed hidden state can be obtained by $\hat{h}_{p,n}^{t-1}=\gamma_{p,n}^t * h_{p,n}^{t-1}$. This factor ensures that the hidden state decays smoothly with longer gaps, naturally reflecting the uncertainty caused by irregular sampling. The decayed state is then combined with the new observation through a gated update mechanism with the sigmoid function $\sigma$, as shown in Eq.~\ref{eq:gate}:
\begin{equation}
\begin{aligned}
r_{p,n}^t 
&= \sigma\!\left(W_r \,[e_{p,n}^t \,\|\, \hat{h}_{p,n}^{t-1}]\right), \\
h_{p,n}^t 
&= (1 - r_{p,n}^t) \cdot \hat{h}_{p,n}^{t-1} 
  + r_{p,n}^t \cdot e_{p,n}^t .
\end{aligned}
\label{eq:gate}
\end{equation}

Notably, the elapsed sampling interval $\Delta t$ is obtained by the previous or next observation timestamp of variable $v$, which is shown as Eq.~\ref{eq:t}:
\begin{equation}
\Delta t_{i,v} =
\begin{cases}
\dfrac{\Delta t_{i,v}^{-}+\Delta t_{i,v}^{+}}{2},
& \text{if } t_{i-1,v},\, t_{i+1,v} \text{ exist}, \\[4pt]
\Delta t_{i,v}^{-} =t_{i,v} - t_{i-1,v},
& \text{if only}\ t_{i-1,v}\ \text{exist}, \\[4pt]
\Delta t_{i,v}^{+} =t_{i+1,v} - t_{i,v},
& \text{if only}\ t_{i+1,v}\ \text{exist}, \\[4pt]
\dfrac{t_{\max}}{2},
& \text{otherwise}.
\end{cases}
\label{eq:t}
\end{equation}



\textbf{Node-specific Updates.} To further leverage the historical states of the variables of a patient, we design a hidden state-aware node-specific attention mechanism that performs node-specific updates before patient nodes enter the graph network. Since there is no hidden state in the first time step, this mechanism is used from the second time step. Specifically, given the previous hidden state of all variables $\textbf{h}^{t-1}$, our state-aware attention output $\mathbf{v'}_p^t$ is computed as Eq.~\ref{eq:SA}:
\begin{equation}
    \mathbf{a}_p^t = \text{softmax}\Big( \frac{\mathbf{v}_{p}^{t}(\mathbf{h}^{t-1})^\top}{\sqrt{d}} \Big) \mathbf{h}^{t-1}, \quad \mathbf{v'}_p^t = W_{\text{proj}}\cdot \mathbf{a}_p^t, 
    \label{eq:SA}
\end{equation}
where the softmax is applied over variable nodes to capture variable-specific temporal context, and the attended feature is then projected to update the patient node embedding. This node-specific attention allows each patient node to selectively incorporate the most relevant information from its historical variable states. By doing so, the patient representation is enriched with fine-grained, variable-wise temporal context, which is particularly important for handling irregularly sampled medical time series and ensuring that subsequent graph message passing operates on a temporally informed patient embedding.

\subsection{Common State Codebook Learning.} Although the bipartite graph construction and temporal decay encoding mechanism effectively capture irregular sampling patterns of IMTS, the learned representations of patient and variable nodes may still be limited in learning the common patterns among different patients. To address this, we introduce a learnable soft-codebook that serves as a set of global prototypes to regularize and align node embeddings.

Formally, let $\mathbf{C} \in \mathbb{R}^{K \times d}$ denote the codebook with $K$ learnable prototypes of dimension $d$. Given the node embeddings $\mathbf{g}_i =[v_{p} || v_{n}] \in \mathbb{R}^d$, we first compute the cosine similarity between each node and all codebook entries. Then the quantized representation can be obtained as a weighted sum over prototypes. This process is illustrated as Eq.~\ref{eq:codebook}:
\begin{equation}
    s_{i,k} = \frac{\mathbf{g}_i}{\|\mathbf{g}_i\|} \cdot \frac{\mathbf{c}_k}{\|\mathbf{c}_k\|},\ \mathbf{c}_k \in \mathbf{C},\  \mathbf{g}_i^{\text{quant}} = \sum_{k=1}^{K} w_{i,k} \cdot \mathbf{c}_k.
\label{eq:codebook}
\end{equation}

Finally, we apply a residual update with an adaptive scaling factor to balance the original embedding and the quantized representation as Eq.~\ref{eq:codefuse}
\begin{equation}
    \mathbf{g}_i \leftarrow \mathbf{g}_i + \alpha_i \cdot \mathbf{g}_i^{\text{quant}}, 
    \quad \alpha_i = \frac{\|\mathbf{g}_i^{\text{quant}}\|}{\|\mathbf{g}_i\| + \epsilon}.
    \label{eq:codefuse}
\end{equation}

This learnable codebook plays two key roles: (1) it compresses noisy continuous representations into a compact set of prototypes, improving robustness to irregular sampling; and (2) it aligns variable dynamics across patients by projecting them onto shared prototypes, thereby enhancing the generalizability of learned patient representations.

\subsection{Training and Inference}
After temporal updating and graph propagation, DBGL produces the patient node embeddings $\mathbf{v} \in \mathbb{R}^{B \times d}$ and the per-batch, per-variable hidden states $\mathbf{h}\in \mathbb{R}^{B \times N \times d}$, where $B$ is the batch size, $N$ is the variable count, and $d$ is the hidden dimension. To emphasize the contribution of observed variables, we first apply a re-weighting operation based on the observation mask, followed by flattening the variable states into a single vector, as shown in Eq.~\ref{eq:his}:
\vspace{-1mm}
\begin{equation}
\begin{aligned}
\mathbf{h} &\leftarrow \mathbf{h} 
+ \mathbf{h} \cdot 
  \text{softmax}\Big(\sum\nolimits_{v} \text{Mask}_{v}\Big), \\
\mathbf{h} &\leftarrow \text{reshape}(\mathbf{h}, [B, -1]),
\end{aligned}
\label{eq:his}
\end{equation}
where $\text{Mask}_{v}$ indicates the observation status of variable $v$. This re-weighting ensures that frequently observed variables have a stronger influence on the patient representation, while the flattening step facilitates integration with subsequent codebook retrieval and classification.  

Additionally, to better enable the codebook itself to capture underlying patient state patterns, we compute the similarity between patient node embeddings $\mathbf{g}_p$ and the learnable codebook $\mathbf{C}$ via cosine similarity and select the most similar codebook entry for each patient node, as Eq.~\ref{eq:code_final}:  
\vspace{-1mm}
\begin{equation}
    \text{Sim} = \frac{\mathbf{g}_p}{\|\mathbf{g}_p\|} \cdot \frac{\mathbf{C}^\top}{\|\mathbf{C}\|}, \quad \mathbf{c}_p = \arg\max_{j} \text{Sim}_{p,j},
    \label{eq:code_final}
\end{equation}
where $\mathbf{c}_p$ denotes the retrieved code vector for patient $p$. The learnable codebook not only provides a mechanism to constrain patient embeddings via retrieval but also evolves into a set of representative latent states during training.

Finally, we concatenate the patient node embedding, its matched code vector, and the flattened hidden representation to form the final representation for classification:  
\begin{equation}
    \mathbf{z}_p = [\mathbf{g}_p ; \mathbf{c}_p ; \mathbf{h}], \quad \hat{\mathbf{y}}_{p} = MLP(\mathbf{z}_p)
    \label{eq:concat}
\end{equation}
which is then fed into a classification layer to predict the clinical outcome $\hat{\mathbf{y}}_{p}$, and Cross Entropy loss is used for optimization.

\section{Experiment}

\subsection{Experimental Setup}
\textbf{Datasets and baselines.} We conduct extensive experiments on four widely used irregular medical time series (IMTS) datasets: P19 \cite{reyna2020early}, PhysioNet \cite{goldberger2000physiobank}, MIMIC-III \cite{johnson2016mimic}, and P12 \cite{silva2012predicting}. Following prior work \cite{shuklamulti}, PhysioNet is considered a reduced version of P12. To comprehensively evaluate the effectiveness of DBGL, we compare DBGL against two categories of baselines: non-graphical methods and graph-based methods. Non-graphical methods include GRU-D \cite{che2018recurrent}, ODE-RNN \cite{rubanova2019latent}, IP-Net \cite{shukla2019interpolation}, SeFT \cite{horn2020set}, mTAND \cite{shukla2021multi}, DGM$^2$-O \cite{wu2021dynamic}, StraTS \cite{tipirneni2022self}, DuETT  \cite{labach2023duett}, ViTST \cite{li2023time} and Warpformer \cite{zhang2023warpformer}. Graph-based methods include MTGNN \cite{wu2020connecting}, Raindrop \cite{zhang2021graph}, ISIM \cite{chen2025integrating}, TimeCHEAT \cite{liu2025timecheat}, and KEDGN \cite{luo2024knowledge}. Details about these datasets and baselines can be found in Appendix~\ref{ap1} and Appendix~\ref{ap2}.

\textbf{Implementation.} We train DBGL with Adam optimizer \cite{kingma2014adam} for 30 epochs, employing an early stopping strategy with a patience of 5 epochs to prevent overfitting. To better verify the robustness of DBGL, we adopt the same setting on all datasets, with a learning rate of 0.005 and a batch size of 256. The hidden state is set as 16, and the codebook size is set as 4096. The layers of EdgeSAGE are set to 2. All experiments are conducted on a NVIDIA GeForce RTX 4090 GPU. We conduct five repeated experiments with different random seeds on each task for evaluation. Since all tasks are binary classification, we adopt AUC-PRC and AUC-ROC as the evaluated metrics.

\subsection{Main Results}
\textbf{IMTS classification.} We first evaluate DBGL on IMTS classification tasks across the four datasets. This setup allows us to compare DBGL against both non-graphical and graph-based baselines under conventional evaluation metrics, highlighting its effectiveness in modeling irregularly sampled clinical data. The results are shown in the Table~\ref{Res1}.

\begin{table*}[t]
\centering
\caption{Performance comparison (AUROC \& AUPRC, \%). The best results are in \textbf{bold} and the second-best results are in \underline{underlined}.}
\vspace{-2mm}
\resizebox{\linewidth}{!}{  
\begin{tabular}{ll|cc|cc|cc|cc}
\toprule
& \multirow{2}{*}{Methods} & \multicolumn{2}{c}{P19} & \multicolumn{2}{c}{Physionet} & \multicolumn{2}{c}{MIMIC-III} & \multicolumn{2}{c}{P12} \\
\cmidrule(lr){3-4} \cmidrule(lr){5-6} \cmidrule(lr){7-8} \cmidrule(lr){9-10}
& & AUROC & AUPRC & AUROC & AUPRC & AUROC & AUPRC & AUROC & AUPRC \\
\midrule
\multirow{11}{*}{Non-Graph} & 
GRU-D       & 88.7 $\pm$ 1.2 & 57.6 $\pm$ 2.3 & 79.1 $\pm$ 6.9 & 42.7 $\pm$ 7.2 & 82.2 $\pm$ 1.8 & 43.3 $\pm$ 2.1 & 79.6 $\pm$ 0.6 & 41.7 $\pm$ 1.8 \\
& ODE-RNN      & 87.1 $\pm$ 1.0 & 52.6 $\pm$ 3.2 & 75.5 $\pm$ 2.8 & 33.7 $\pm$ 4.1 & 81.0 $\pm$ 0.6 & 42.3 $\pm$ 0.7 & 78.8 $\pm$ 0.6 & 37.4 $\pm$ 2.6 \\
& IP-Net       & 90.2 $\pm$ 0.2 & 58.6 $\pm$ 0.8 & 86.8 $\pm$ 0.6 & 55.8 $\pm$ 1.4 & 84.1 $\pm$ 0.1 & 47.1 $\pm$ 0.9 & 83.7 $\pm$ 0.3 & 46.3 $\pm$ 1.3 \\
& SeFT         & 84.0 $\pm$ 0.3 & 49.3 $\pm$ 0.5 & 75.5 $\pm$ 0.2 & 29.4 $\pm$ 0.9 & 67.9 $\pm$ 0.2 & 23.2 $\pm$ 0.4 & 78.1 $\pm$ 0.5 & 35.9 $\pm$ 0.8 \\
& mTAND        & 82.9 $\pm$ 0.9 & 32.3 $\pm$ 1.5 & 86.9 $\pm$ 1.3 & 52.5 $\pm$ 1.3 & 83.8 $\pm$ 0.3 & 46.6 $\pm$ 0.5 & 85.3 $\pm$ 0.3 & 49.3 $\pm$ 1.0 \\
& $\text{DGM}^{2}$-O       & 91.6 $\pm$ 0.5 & 60.0 $\pm$ 1.3 & 85.8 $\pm$ 0.7 & 50.4 $\pm$ 3.2 & 81.0 $\pm$ 0.9 & 37.6 $\pm$ 1.1 & 85.8 $\pm$ 0.1 & 48.3 $\pm$ 0.7 \\

& StraTS       & 91.2 $\pm$ 0.3 & 58.4 $\pm$ 1.4 & 84.9 $\pm$ 1.5 & 47.3 $\pm$ 5.3 & 84.4 $\pm$ 0.4 & 46.4 $\pm$ 0.8 & 86.7 $\pm$ 0.7 & 52.1 $\pm$ 1.5 \\
& DuETT        & 88.2 $\pm$ 0.5 & 56.0 $\pm$ 3.9 & 81.3 $\pm$ 1.4 & 44.9 $\pm$ 1.4 & 78.8 $\pm$ 0.8 & 34.3 $\pm$ 1.0 & 83.4 $\pm$ 1.2 & 45.4 $\pm$ 1.5 \\
& ViTST        & 91.7 $\pm$ 0.1 & 57.5 $\pm$ 0.7 & 81.3 $\pm$ 1.9 & 37.4 $\pm$ 2.9 & 81.8 $\pm$ 0.3 & 39.6 $\pm$ 1.3 & 86.3 $\pm$ 0.1 & 50.8 $\pm$ 1.5 \\
& Warpformer   & 91.8 $\pm$ 0.4 & 60.6 $\pm$ 2.6 & 83.3 $\pm$ 0.7 & 43.5 $\pm$ 2.3 & 84.6 $\pm$ 0.5 & 47.4 $\pm$ 0.9 & 85.4 $\pm$ 0.5 & 50.4 $\pm$ 1.5 \\
\midrule
\multirow{4}{*}{Graph-based} 
& MTGNN        & 88.5 $\pm$ 1.0 & 55.8 $\pm$ 1.5 & 77.1 $\pm$ 4.4 & 35.4 $\pm$ 7.3 & 78.5 $\pm$ 2.3 & 35.2 $\pm$ 3.1 & 82.1 $\pm$ 1.5 & 41.8 $\pm$ 2.1 \\
& Raindrop     & 89.4 $\pm$ 0.6 & 61.2 $\pm$ 1.1 & 82.7 $\pm$ 1.4 & 41.2 $\pm$ 3.6 & 79.8 $\pm$ 1.3 & 35.2 $\pm$ 1.1 & 82.2 $\pm$ 1.1 & 43.3 $\pm$ 2.1 \\
& ISIM & 91.6 $\pm$ 0.9 & 59.6 $\pm$ 1.3 & - & -& -& -& 86.0 $\pm$ 0.3 & 50.4 $\pm$ 2.1 \\
& TimeCHEAT & 89.5 $\pm$ 1.9 & 56.1 $\pm$  4.6  & - & -& -& -& 84.5 $\pm$ 0.7 & 48.2 $\pm$ 1.9 \\
& KEDGN-Name   & \underline{92.3 $\pm$ 1.0} & \underline{62.5 $\pm$ 0.7} & 87.9 $\pm$ 1.4 & 56.0 $\pm$ 3.2 & 84.8 $\pm$ 0.3 & \underline{48.4 $\pm$ 1.5} & 87.1 $\pm$ 0.8 & 54.1 $\pm$ 2.6 \\
& KEDGN-Wiki   & 92.2 $\pm$ 0.6 & 62.3 $\pm$ 1.4 & \underline{88.2 $\pm$ 1.1} & \underline{57.5 $\pm$ 2.5} & 84.3 $\pm$ 0.9 & 47.7 $\pm$ 1.8 & 87.0 $\pm$ 0.3 & 53.1 $\pm$ 0.5 \\
& KEDGN-ChatGPT& 92.2 $\pm$ 0.5 & 62.0 $\pm$ 1.3 & 87.9 $\pm$ 0.2 & 57.1 $\pm$ 1.8 & 85.1 $\pm$ 0.3 & 48.3 $\pm$ 1.6 & \underline{87.8 $\pm$ 0.5} & \underline{54.5 $\pm$ 1.5} \\
\midrule
\rowcolor[gray]{0.95} \multirow{2}{*}{Ours} & DBGL         & \textbf{93.3 $\pm$ 0.5} & \textbf{66.3 $\pm$ 1.4} & \textbf{89.1 $\pm$ 0.3} & \textbf{60.8 $\pm$ 2.2} & \textbf{85.2 $\pm$ 0.4} & \textbf{50.2 $\pm$ 1.0} & \textbf{88.1 $\pm$ 0.4} & \textbf{56.3 $\pm$ 1.0} \\
\rowcolor[gray]{0.95} & Gain         & \textcolor{red}{\textbf{+1.0}} & \textcolor{red}{\textbf{+3.8}} & \textcolor{red}{\textbf{+0.9}} & \textcolor{red}{\textbf{+3.6}} & \textcolor{red}{\textbf{+0.1}} & \textcolor{red}{\textbf{+1.9}} & \textcolor{red}{\textbf{+0.3}} & \textcolor{red}{\textbf{+1.8}} \\
\bottomrule
\end{tabular}
}
\label{Res1}
\end{table*}

From the results, several key observations emerge. First, DBGL consistently outperforms all baselines across datasets, achieving the highest AUROC and AUPRC in every case. Compared to the strongest graph-based competitor (KEDGN variants), DBGL yields improvements of up to +3.8\% in AUPRC and +1.0\% in AUROC, demonstrating its superior ability to capture variable dependencies and temporal irregularities. Second, while non-graphical methods such as GRU-D, ODE-RNN, and IP-Net benefit from sequential modeling, they generally lag behind graph-based approaches, highlighting the importance of explicitly modeling inter-variable relations in clinical time series. Third, existing graph-based methods, including MTGNN, Raindrop, and KEDGN variants, achieve strong performance by capturing spatial-temporal dependencies, yet they still fall short of DBGL, likely due to their limited handling of irregular sampling and variable-specific decay patterns.  

Overall, these results validate that the bipartite graph construction in DBGL, together with its node-specific decay mechanism, effectively models the inherent irregularities in medical time series, leading to more accurate and robust representations for downstream prediction tasks.

\textbf{Leave-variables-out.}
To better verify the robustness of DBGL, we also conduct the Leave-variables-out experiments as in previous work \cite{luo2024knowledge}, which test the model when a subset of variables is completely missing. The discarding rate of each variable is set from 10\% to 50\% with a step of 10\%, and all their observations in both validation and test sets will also be hidden. The results, compared with baselines onthe  P12 dataset, are shown in Table~\ref{res2}.

\begin{table*}[t]
\centering
\caption{Performance comparison (AUROC \& AUPRC, \%) under different variable discard ratios on P12 dataset. The best results are highlighted in \textbf{bold} and the second-best results are in \underline{underlined}.}
\vspace{-2mm}
\resizebox{\linewidth}{!}{
\begin{tabular}{l|cc|cc|cc|cc|cc}
\toprule
\multirow{2}{*}{Method} & 
\multicolumn{2}{c}{10\%} & 
\multicolumn{2}{c}{20\%} & 
\multicolumn{2}{c}{30\%} & 
\multicolumn{2}{c}{40\%} & 
\multicolumn{2}{c}{50\%} \\
\cmidrule(lr){2-3} \cmidrule(lr){4-5} \cmidrule(lr){6-7} \cmidrule(lr){8-9} \cmidrule(lr){10-11}
& AUROC & AUPRC & AUROC & AUPRC & AUROC & AUPRC & AUROC & AUPRC & AUROC & AUPRC \\
\midrule
GRU-D & 68.6 $\pm$ 2.3 & 35.8 $\pm$ 2.2 & 68.2 $\pm$ 2.1 & 34.5 $\pm$ 2.9 & 66.8 $\pm$ 3.3 & 32.7 $\pm$ 4.6 & 65.8 $\pm$ 4.0 & 31.3 $\pm$ 5.2 & 65.1 $\pm$ 4.1 & 30.4 $\pm$ 5.5 \\
mTAND & 74.9 $\pm$ 0.6 & 37.7 $\pm$ 0.6 & 74.0 $\pm$ 1.3 & 36.5 $\pm$ 1.5 & 71.4 $\pm$ 3.8 & 34.1 $\pm$ 3.7 & 70.6 $\pm$ 3.6 & 33.2 $\pm$ 3.7 & 70.1 $\pm$ 3.5 & 32.5 $\pm$ 3.6 \\
$\text{DGM}^{2}$-O & 76.3 $\pm$ 1.1 & 39.3 $\pm$ 1.5 & 76.1 $\pm$ 1.1 & 38.2 $\pm$ 1.7 & 74.8 $\pm$ 2.2 & 36.8 $\pm$ 2.6 & 72.0 $\pm$ 5.3 & 34.3 $\pm$ 5.0 & 70.4 $\pm$ 5.9 & 32.7 $\pm$ 5.7 \\
MTGNN & 71.2 $\pm$ 2.1 & 30.5 $\pm$ 1.5 & 70.3 $\pm$ 3.3 & 29.7 $\pm$ 2.8 & 68.9 $\pm$ 4.2 & 28.5 $\pm$ 3.3 & 68.1 $\pm$ 4.7 & 27.7 $\pm$ 3.6 & 67.6 $\pm$ 5.2 & 27.2 $\pm$ 3.8 \\
Raindrop & 73.2 $\pm$ 1.6 & 32.4 $\pm$ 0.9 & 73.0 $\pm$ 1.6 & 31.7 $\pm$ 1.1 & 72.2 $\pm$ 2.6 & 31.1 $\pm$ 2.7 & 71.5 $\pm$ 3.5 & 30.6 $\pm$ 3.5 & 70.8 $\pm$ 4.2 & 29.7 $\pm$ 4.3 \\
DuETT & 73.9 $\pm$ 1.7 & 35.8 $\pm$ 2.3 & 74.7 $\pm$ 1.8 & 35.3 $\pm$ 2.0 & 73.6 $\pm$ 2.2 & 34.1 $\pm$ 2.4 & 72.8 $\pm$ 2.6 & 33.3 $\pm$ 2.7 & 72.3 $\pm$ 2.7 & 32.6 $\pm$ 2.8 \\
Warpformer & 75.9 $\pm$ 0.7 & 37.3 $\pm$ 2.2 & 75.6 $\pm$ 0.8 & 36.7 $\pm$ 2.3 & 73.8 $\pm$ 2.9 & 34.3 $\pm$ 4.1 & 72.8 $\pm$ 3.4 & 33.0 $\pm$ 4.6 & 72.1 $\pm$ 3.7 & 32.2 $\pm$ 4.7 \\
KEDGN & \underline{79.7 $\pm$ 0.4} & \underline{43.6 $\pm$ 1.2} & \underline{79.2 $\pm$ 0.8} & \underline{42.5 $\pm$ 1.6} & \underline{77.7 $\pm$ 2.2} & \underline{40.0 $\pm$ 4.0} & \underline{77.2 $\pm$ 2.2} & \underline{39.6 $\pm$ 3.7} & \underline{76.9 $\pm$ 2.2} & \underline{39.2 $\pm$ 3.5} \\
\midrule
\rowcolor[gray]{0.95} DBGL & \textbf{86.3 $\pm$ 0.8} & \textbf{52.1 $\pm$ 1.9} & \textbf{86.0 $\pm$ 0.7} & \textbf{51.6 $\pm$ 2.2} & \textbf{83.9 $\pm$ 1.2} & \textbf{48.4 $\pm$ 2.1} & \textbf{83.5 $\pm$ 1.6} & \textbf{47.3 $\pm$ 3.7} & \textbf{81.3 $\pm$ 0.8} & \textbf{42.7 $\pm$ 1.4} \\
\rowcolor[gray]{0.95} Gain       &   \textcolor{red}{\textbf{+6.6}}         &   \textcolor{red}{\textbf{+8.5}}           &  \textcolor{red}{\textbf{+6.8}}            &   \textcolor{red}{\textbf{+9.1}}           &      \textcolor{red}{\textbf{+6.2}}        &          \textcolor{red}{\textbf{+8.4}}    &     \textcolor{red}{\textbf{+6.3}}         &    \textcolor{red}{\textbf{+7.7}}          &    \textcolor{red}{\textbf{+4.4}}          &  \textcolor{red}{\textbf{+3.5}}            \\
\bottomrule
\end{tabular}}
\label{res2}
\end{table*}

DBGL consistently outperforms all competitors across all discard rates. Even when up to 50\% of variables are missing, DBGL maintains strong predictive performance (AUROC 81.3\%, AUPRC 42.7\%), substantially higher than the best baseline (KEDGN) by +4.4\% AUROC and +3.5\% AUPRC at the highest discard level. The performance gap increases as more variables are discarded, indicating that DBGL’s bipartite graph structure and node-specific decay mechanism effectively propagate information from available variables and mitigate information loss. In contrast, both sequential (GRU-D, mTAND) and existing graph-based methods degrade significantly under high missing rates, highlighting the limitations of their imputation or message-passing strategies in handling variable sparsity. More results on other datasets can be found in the Appendix~\ref{ap:31}

Overall, these results demonstrate that DBGL is highly robust to missing variables and can reliably model irregular clinical time series even under substantial data scarcity.

\subsection{Analysis}

We also conduct an analysis of confidence in detecting positive cases from our DBGL and the best baseline KEDGN by examining the average predicted probability on positive-class samples. This metric provides a direct measure of how strongly a model supports its positive predictions, with higher values indicating stronger discriminative capability.  

\begin{figure}[htbp]
  \centering
\includegraphics[width=0.35\textwidth]{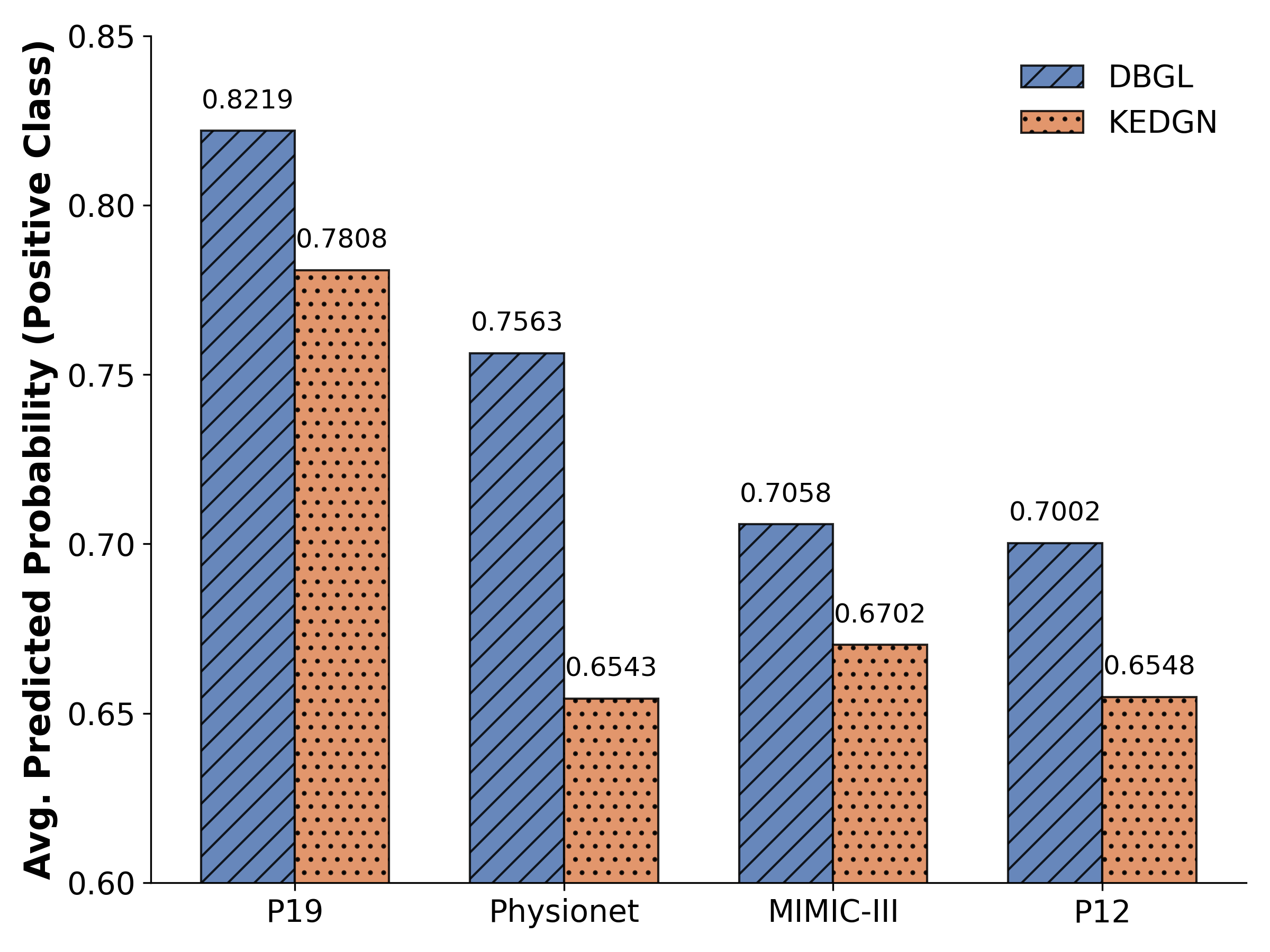}
\vspace{-3mm}
  \caption{Comparison of Positive-class Predicted Probabilities.}
  \label{fig:prob}
\end{figure}

As shown in Figure~\ref{fig:prob}, DBGL consistently yields higher positive-class probabilities than KEDGN across all four datasets. The most notable gain appears on Physionet (0.7563 vs. 0.6543, +0.1020), while consistent improvements are also observed on P19 (+0.0411), MIMIC-III (+0.0356), and P12 (+0.0454). These results suggest that DBGL learns more robust representations, enabling reliable identification of positive samples even under heterogeneous data distributions, which is crucial for the clinical field.  

From a clinical perspective, stronger confidence in positive cases reflects more reliable disease detection, which is critical for reducing missed diagnoses and supporting decision-making in time-critical, real-world settings such as the ICU. For example, in P19, accurate identification of patients at risk of sepsis within 6 hours enables timely interventions, while in P12, detecting patients likely to experience prolonged ICU stays informs resource allocation during the first 48 hours. Similarly, in MIMIC-III and Physionet, high-confidence predictions of in-hospital mortality allow clinicians to prioritize monitoring and treatment for high-risk patients. Across these diverse datasets, models that exhibit strong positive-case confidence provide more trustworthy and actionable clinical insights.

\subsection{Ablation Study}
We conduct ablation studies to evaluate the contribution of each key component of DBGL across all four datasets. Specifically, we design the following variants: (1) \textbf{w/o TDE}: removing the temporal decay encoding mechanism; (2) \textbf{w/o SNA}: removing the state-aware node-specific attention for patient node updates; (3) \textbf{w/o HVS}: discarding hidden variable states from the final representation; (4) \textbf{w/o CB}: removing the learnable codebook; (5) \textbf{w/o MCV}: discarding the matched code vector retrieved from the codebook for classification. (6) \textbf{w/o TE}: removing the Time Embedding. These variants enable us to disentangle the effects of temporal modeling, node-specific attention, hidden state preservation, and codebook constraints on the overall performance. 

\begin{table*}[t]
\centering
\caption{Ablation study results (AUROC \& AUPRC, \%) on four clinical datasets. The best results are highlighted in 
\vspace{-2mm}
\textbf{bold}.}
\scalebox{0.775}{
\begin{tabular}{l|cc|cc|cc|cc}
\toprule
Method & \multicolumn{2}{c}{P19} & \multicolumn{2}{c}{Physionet} & \multicolumn{2}{c}{MIMIC-III} & \multicolumn{2}{c}{P12} \\
\cmidrule(lr){2-3} \cmidrule(lr){4-5} \cmidrule(lr){6-7} \cmidrule(lr){8-9}
& AUROC & AUPRC & AUROC & AUPRC & AUROC & AUPRC & AUROC & AUPRC \\
\midrule
w/o TDE & 92.9 $\pm$ 0.7 & 64.3 $\pm$ 2.0 & 88.7 $\pm$ 0.8 & 59.1 $\pm$ 2.2 & 84.6 $\pm$ 1.1 & 48.2 $\pm$ 2.4 & 87.2 $\pm$ 0.5 & 53.3 $\pm$ 1.9 \\
w/o SNA & 92.4 $\pm$ 0.6 & 65.0 $\pm$ 1.0 & 88.2 $\pm$ 1.1 & 56.7 $\pm$ 5.1 & 84.9 $\pm$ 0.3 & 49.1 $\pm$ 1.0 & 87.3 $\pm$ 0.6 & 53.3 $\pm$ 2.3 \\
w/o HVS & 92.3 $\pm$ 0.5 & 64.9 $\pm$ 1.2 & 87.1 $\pm$ 2.3 & 55.6 $\pm$ 3.8 & 84.5 $\pm$ 1.8 & 46.9 $\pm$ 4.1   & 86.9 $\pm$ 0.5 & 53.1 $\pm$ 0.8 \\
w/o CB  & 92.8 $\pm$ 0.5 & 65.3 $\pm$ 1.4 & 88.5 $\pm$ 1.4 & 58.0 $\pm$ 4.8 & 84.9 $\pm$ 0.3 & 49.5 $\pm$ 0.9 & 87.7 $\pm$ 0.5 & 54.2 $\pm$ 1.7 \\
w/o MCV & 93.1 $\pm$ 0.3 & 66.1 $\pm$ 1.4 & 88.9 $\pm$ 0.5 & 59.5 $\pm$ 1.1 & 85.1 $\pm$ 0.2 & 50.0 $\pm$ 0.2 & 87.9 $\pm$ 0.2 & 54.6 $\pm$ 0.5\\
w/o TE & 86.8 $\pm$ 2.0   & 63.7 $\pm$ 1.7   & 88.2 $\pm$ 0.8      & 58.8 $\pm$ 2.4      & 85.0 $\pm$ 0.3      & 50.1 $\pm$ 0.3      & 87.1 $\pm$ 0.5  & 53.8 $\pm$ 1.9 \\
\midrule
\rowcolor[gray]{0.95} Full    & \textbf{93.3 $\pm$ 0.5} & \textbf{66.3 $\pm$ 1.4} & \textbf{89.1 $\pm$ 0.3} & \textbf{60.8 $\pm$ 2.2} & \textbf{85.2 $\pm$ 0.4} & \textbf{50.2 $\pm$ 1.0} & \textbf{88.1 $\pm$ 0.4} & \textbf{56.3 $\pm$ 1.0} \\
\bottomrule
\end{tabular}
}
\label{ablation_table}
\end{table*}

\begin{table*}[t]
\centering
\caption{Performance comparison of different kernel types (AUROC \& AUPRC, \%). The best results are highlighted in \textbf{bold}.}
\vspace{-2mm}
\scalebox{0.775}{
\begin{tabular}{l|cc|cc|cc|cc}
\toprule
Kernel & \multicolumn{2}{c}{P19} & \multicolumn{2}{c}{Physionet} & \multicolumn{2}{c}{MIMIC-III} & \multicolumn{2}{c}{P12} \\
\cmidrule(lr){2-3} \cmidrule(lr){4-5} \cmidrule(lr){6-7} \cmidrule(lr){8-9}
& AUROC & AUPRC & AUROC & AUPRC & AUROC & AUPRC & AUROC & AUPRC \\
\midrule
Exp            & 92.9 $\pm$ 0.3 & 65.2 $\pm$ 0.6 & 88.8 $\pm$ 0.7 & 58.2 $\pm$ 1.9 & 84.9 $\pm$ 0.2 & 49.9 $\pm$ 0.5 & 87.0 $\pm$ 0.3 & 53.0 $\pm$ 1.0 \\
MLP + Linear   & 93.2 $\pm$ 0.7 & 65.6 $\pm$ 0.9 & \textbf{89.3 $\pm$ 0.4} & 60.8 $\pm$ 2.2 & 84.8 $\pm$ 0.2 & 49.5 $\pm$ 1.0 & 87.6 $\pm$ 0.3 & 54.2 $\pm$ 0.5 \\
MLP + Gaussian & 93.3 $\pm$ 0.4 & 66.1 $\pm$ 1.6 & 89.3 $\pm$ 0.4 & 60.7 $\pm$ 2.4 & 84.6 $\pm$ 0.9 & 48.3 $\pm$ 2.5 & 87.7 $\pm$ 0.5 & 54.5 $\pm$ 1.3 \\
\rowcolor[gray]{0.95} MLP + Exp      & \textbf{93.3 $\pm$ 0.5} & \textbf{66.3 $\pm$ 1.4} & 89.1 $\pm$ 0.3 & \textbf{60.8 $\pm$ 2.2} & \textbf{85.2 $\pm$ 0.4} & \textbf{50.2 $\pm$ 1.0} & \textbf{88.1 $\pm$ 0.4} & \textbf{56.3 $\pm$ 1.0} \\
\bottomrule
\end{tabular}
}
\label{kernel_table}
\end{table*}

Table~\ref{ablation_table} reports the ablation results. Several observations can be made. First, removing the temporal decay encoding mechanism (\textbf{w/o TDE}) consistently leads to a noticeable drop in both AUROC and AUPRC across datasets, highlighting the importance of explicitly modeling temporal decay in irregular sequences. Second, discarding the state-aware node-specific attention (\textbf{w/o SNA}) results in degraded performance, especially on Physionet, where the AUPRC drops sharply, confirming that patient-specific node updates are crucial for handling heterogeneous and irregular dynamics. Third, excluding hidden variable states (\textbf{w/o HVS}) yields the most pronounced performance decline, particularly in AUPRC, indicating that preserving latent states is essential for capturing long-term dependencies under irregular sampling. Fourth, removing the codebook (\textbf{w/o CB}) also impairs performance, though to a lesser extent, suggesting that the learned representation space benefits from codebook constraints. Finally, ignoring the matched code vector (\textbf{w/o MCV}) reduces both AUROC and AUPRC compared to the full model, showing the utility of codebook-guided classification. Overall, the full DBGL model achieves the best performance on all datasets, demonstrating the complementary contributions of each component.

\textbf{Kernel funcition ablation in TDE.} To investigate the effect of alternative continuous-time kernels, we conducted additional experiments by replacing the original exponential decay: \textbf{Gaussian Kernel}: $\gamma=exp(-(\lambda \cdot \Delta t)^{2})$, and \textbf{Linear Kernel}: $\gamma = max(1-\lambda \cdot \Delta t, 0)$., where $\lambda$ is the decay rate. Both of these keep the MLP-learned decay rate. Experimental results are shown in Table~\ref{kernel_table}.

The MLP + Exp kernel used in DBGL achieves the best performance, confirming the benefit of learning variable-specific decay rates. Alternative kernels perform slightly worse, demonstrating DBGL’s robustness to kernel choice. Softplus function ensures $\lambda \geq 0$, so $e^{(-\lambda * \Delta t)} \in (0,1]$, preventing overflow/underflow even for large $\Delta t$. $\gamma$ decreases monotonically with $\Delta t$ and varies with variable-specific edge attributes via the MLP, mapping variables with different temporal dynamics to distinct decay behaviors. This ensures both smooth gradients and extraction of variable dynamics. In summary, the combination of Softplus and the exponential function guarantees a smooth, differentiable, and numerically stable decay, while effectively capturing heterogeneous temporal dynamics across variables.

\subsection{Comparison to Patient-Timestep Graph.}

\begin{figure}[htbp]
  \centering
\includegraphics[width=0.35\textwidth]{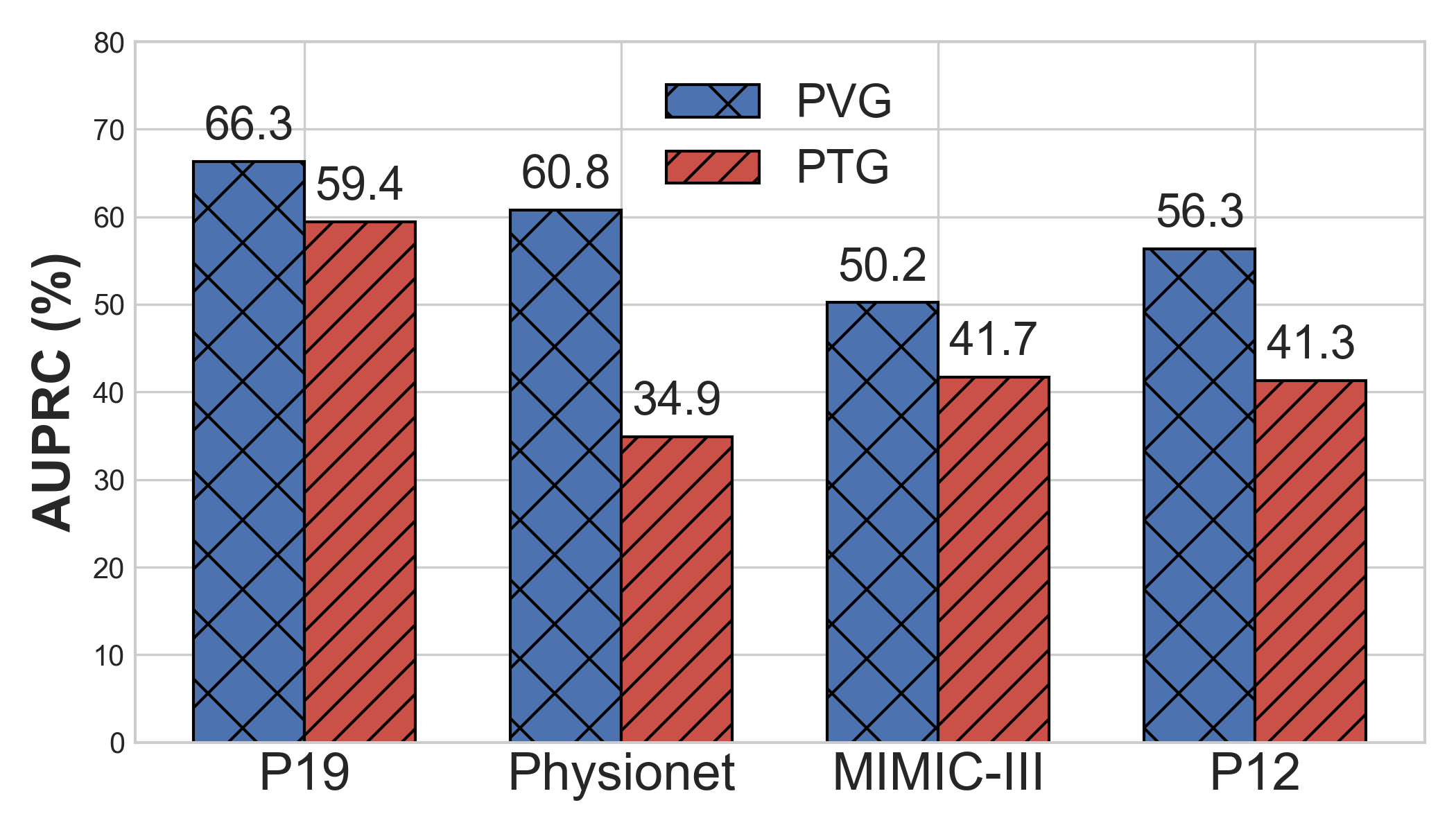}
\vspace{-3mm}
  \caption{Comparison of graph construction.}
  \label{fig:graph}
\end{figure}

To better verify the effect of the proposed patient-variable bipartite graph, we also conduct a variant of DBGL that constructs each variable as a patient-timestep bipartite graph. The motivation behind this comparison is that the patient-timestep bipartite graph, where the emphasis is placed on temporal alignment across different measurements. The experimental results on four datasets are shown in Figure~\ref{fig:graph}. We regard our DBGL as PVG (patient-variable graph) and the patient-timestep graph as PTG. The experimental results show that PTG has experienced an extremely significant performance degradation on each dataset.   

The advantage of the patient-timestep bipartite graph lies in its intuitive depiction of patients’ changes over time, emphasizing temporal alignment and dynamic trends, with a simple structure that is well-suited for modeling global temporal dependencies. However, this design may overlook the heterogeneous nature of variables and their irregular sampling characteristics, since it treats each variable instance at a given time as independent nodes without explicitly encoding variable-level semantics. By contrasting the two designs, we aim to highlight the advantage of modeling variable-level relationships under irregular sampling, which allows DBGL to better capture variable-specific dynamics and richer cross-variable dependencies. This comparison shows the necessity of our proposed DBGL and clarifies its contribution beyond existing timestep-centered formulations.

\section{Conclusion}

In this paper, we propose DBGL, a novel framework for IMTS classification. DBGL directly encodes the irregular sampling pattern of each variable in each time step into a patient-variable bipartite graph, effectively capturing temporal sampling irregularities. Using this graph, we introduce a temporal decay encoding mechanism, which captures each variable's decay rate proportional to its sampling interval and observation. This design enables DBGL to utilize the unique historical state, sampling interval, and decay dynamics of each variable to perform precise state updates at every time step, resulting in more informative representations. Extensive experiments on four publicly available clinical datasets demonstrate that DBGL consistently outperforms state-of-the-art baselines. Overall, DBGL provides a general and flexible framework for modeling irregularly sampled clinical time series, offering both improved predictive performance. In future work, we plan to extend DBGL to handle multimodal patient data, incorporate uncertainty estimation for clinical decision support, and explore its application to other time-critical healthcare tasks.


\bibliography{example_paper}
\bibliographystyle{icml2026}

\newpage
\appendix
\onecolumn
\section*{Appendix}
\section{Dataset details}
\label{ap1}

Four public irregular medical time series datasets are used for evaluation in this work. The statistics of these datasets are summarized in Table~\ref{tab:dataset}. 

\begin{table}[h]
    \centering
    \caption{Dataset statistics}
    \scalebox{1.0}{
    \begin{tabular}{cccccc}
    \toprule
        Datasets & \# Samples &  \# Variables & \# Max Length & Missing Ratio & Task  \\
        \midrule
        P19 & 38803 & 34 & 60 & 94.9\% & Sepsis Prediction \\
        P12 & 11988 & 36 & 215 & 88.4\% & Stay Predcition\\
        MIMIC-III & 21107 & 16 & 292 & 65.5\% & Mortality Prediction \\
        Physionet & 3997 & 36 & 215 & 84.9\%  & Mortality Prediction\\
        \bottomrule
    \end{tabular}
    }
    \label{tab:dataset}
\end{table}

\textbf{P19 Dataset}. Originating from the PhysioNet 2019 Sepsis Early Prediction Challenge \cite{reyna2020early}, this dataset comprises medical records of 38,803 patients. All samples are associated with a binary classification label indicating whether sepsis occurred within the subsequent 6 hours. Each record includes 34 time-series variables (with a maximum time span of 60 hours) and a static feature vector containing attributes such as age, gender, time from hospital admission to ICU admission, ICU type, and length of stay in days. Samples with abnormal time-series lengths were excluded following the criteria outlined in \cite{zhanggraph}. The dataset is available at: https://physionet.org/content/challenge-2019/1.0.0/. This dataset performs patient-level data splitting, wtih 8:1:1 for training, validation, and test set.

\textbf{P12.} After removing 12 invalid samples identified in \cite{horn2020set}, the P12 dataset [46] contains 11,988 valid patient records. The binary labels are determined based on the length of ICU stay: a negative label indicates a short stay ($\leq$ 3 days), and a positive label indicates a long stay ($>$ 3 days). Each patient’s record includes multivariate time series collected from 36 types of sensors (excluding weight measurements) during the first 48 hours of ICU admission, along with a 9-dimensional static feature vector containing demographic and other relevant attributes. The data can be accessed at: https://physionet.org/content/challenge-2012/1.0.0/. This dataset performs patient-level data splitting, wtih 8:1:1 for training, validation, and test set.

\textbf{MIMIC-III.} MIMIC-III \cite{johnson2016mimic} is a widely used public medical database containing de-identified electronic health records of ICU patients at the Beth Israel Deaconess Medical Center between 2001 and 2012. This study employs its binary in-hospital mortality prediction task for evaluation. The original data includes approximately 57,000 ICU records, covering multidimensional variables such as medication records, in-hospital mortality, and vital signs. After preprocessing, 21,107 samples were obtained, each containing 16 clinical features over 48 hours. The dataset is available at: https://physionet.org/content/mimiciii/1.4/. This dataset performs record-level data splitting, wtih 70\%/15\%/15\% for training, validation, and test set.

\textbf{Physionet.} Physionet \cite{silva2012predicting} contains monitoring data from the first 48 hours of ICU patient admissions. This study primarily uses the in-hospital mortality prediction task. Following the same preprocessing pipeline as applied to the P12 dataset, 3,997 annotated samples were obtained. The data can be found at: https://physionet.org/content/challenge-2012/. This dataset performs patient-level data splitting, wtih 8:1:1 for training, validation, and test set.

\section{Variable Decay Analysis}
\label{vda}
We conducted a systematic analysis of the decay rates of different variables across all datasets. First, we clarify the meaning of the variable decay rate: one of the core components of our method is the “variable decay irregularity” model, which characterizes the temporal dynamics of each clinical variable—specifically, the rate at which its autocorrelation decays over time. A higher decay rate indicates stronger short-term fluctuations (e.g., heart rate and blood pressure, which typically vary on a minute-level scale), whereas a lower decay rate reflects greater temporal stability (e.g., creatinine and hemoglobin, which often change over hours or days). This metric therefore provides a unified quantitative way to capture clinically meaningful differences in temporal behavior across variables.

To further deal with the concern regarding whether different clinical variables exhibit heterogeneous temporal decay, we conducted an autocorrelation-based analysis on all training samples. For each variable, we computed the empirical autocorrelation over time lags $(\Delta t)$ and fitted an exponential model $ \text{Corr}(\Delta t) = e^{-\lambda \Delta t} $ to estimate its decay rate.

\begin{table*}[h]
\centering
\caption{Variable-specific decay rates ($\lambda$) on the P12 dataset. Higher $\lambda$ indicates faster-changing.}
\begin{tabular}{l|c|l|c|l|c|l|c}
\toprule
Variable & $\lambda$ & Variable & $\lambda$ & Variable & $\lambda$ & Variable & $\lambda$ \\
\midrule
ALP        & 14.7113 & FiO2       & 0.2458 & NISysABP  & 6.0698 & ALT        & 14.6235 \\
GCS        & 1.0948  & Na         & 0.0420 & AST       & 13.8853 & Glucose    & 0.0493 \\
PaCO2      & 0.0225  & Albumin    & 0.0853 & HCO3      & 0.1094 & PaO2       & 0.0280 \\
BUN        & 0.0854  & HCT        & 0.0379 & Platelets & 0.0505 & Bilirubin  & 0.8760 \\
HR         & 1.7301  & RespRate   & 4.1157 & Cholesterol & 7.0000 & K          & 0.0368 \\
SaO2       & 0.0408  & Creatinine & 0.1399 & Lactate   & 13.0000 & SysABP     & 0.1403 \\
DiasABP    & 0.0232  & MAP        & 0.0913 & Temp      & 0.0265 & MechVent   & 9.4304 \\
Mg         & 0.1023  & TropI      & 0.0684 & NIDiasABP & 0.0272 & NIMAP      & 0.9093 \\
TropT      & 14.0000 & Urine      & 0.0986 & WBC       & 0.0545 & pH         & 10.8730 \\
\bottomrule
\end{tabular}
\label{tab:VDAP12}
\end{table*}

\begin{table*}[t]
\centering
\caption{Variable-specific decay rates ($\lambda$) on the P19 dataset. Higher $\lambda$ indicates faster-changing.}
\begin{tabular}{l|c|l|c|l|c|l|c}
\toprule
Variable & $\lambda$ & Variable & $\lambda$ & Variable & $\lambda$ & Variable & $\lambda$ \\
\midrule
Heart rate      & 0.0759  & Temp           & 0.1131  & BaseExcess      & 14.0    & SpO2            & 0.0822  \\
SBP             & 0.0813  & HCO3           & 0.0900  & MAP             & 0.0820  & DBP             & 0.0798  \\
FiO2            & 0.4244  & Resp           & 0.0818  & EtCO2           & 0.1071  & pH              & 0.0953  \\
PaCO2           & 0.7904  & SaO2           & 16.0    & AST             & 16.235  & BUN             & 16.0    \\
Alkalinephos    & 15.0    & Calcium        & 0.1144  & Chloride        & 0.0782  & Creatinine      & 0.1134  \\
Bilirubin\_direct & 15.0   & Glucose        & 0.1376  & Lactate         & 3.7805  & Magnesium       & 0.0850  \\
Phosphate       & 0.1111 & Potassium      & 0.1036  & Bilirubin\_total & 16.0    & TroponinI       & 0.1478  \\
Hct             & 0.1004 & Hgb            & 0.0914  & PTT             & 0.1305  & WBC             & 0.1015  \\
Fibrinogen      & 0.0890 & Platelets      & 0.0896  & $-$             & $-$       & $-$             & $-$       \\
\bottomrule
\end{tabular}
\label{tab:p19_lambda}
\end{table*}

\begin{table*}[t]
\centering
\caption{Variable-specific decay rates ($\lambda$) on the Physionet dataset. Higher $\lambda$ indicates faster-changing.}
\begin{tabular}{l|c|l|c|l|c|l|c}
\toprule
Variable & $\lambda$ & Variable & $\lambda$ & Variable & $\lambda$ & Variable & $\lambda$ \\
\midrule
ALP          & 16.5173 & ALT          & 14.4846 & AST           & 16.0     & Albumin       & 0.1164  \\
BUN          & 0.0400  & Bilirubin    & 14.0    & Cholesterol   & 2.0      & Creatinine    & 0.0502  \\
DiasABP      & 0.0212  & FiO2         & 0.3273  & GCS           & 13.5627  & Glucose       & 0.0586  \\
HCO3         & 0.0791  & HCT          & 0.0392  & HR            & 12.0     & K             & 0.0429  \\
Lactate      & 12.8876 & MAP          & 0.0371  & MechVent      & 9.4304   & Mg            & 14.0    \\
NIDiasABP    & 0.0272  & NIMAP        & 1.0679  & NISysABP      & 11.0     & Na            & 0.0441  \\
PaCO2        & 0.0289  & PaO2         & 0.0326  & Platelets     & 0.0721   & RespRate      & 3.3877  \\
SaO2         & 0.2059  & SysABP       & 0.1289  & Temp          & 0.0435   & TropI         & 0.1035  \\
TropT        & 14.0    & Urine        & 0.0279  & WBC           & 0.1009   & pH            & 11.9039 \\
\bottomrule
\end{tabular}
\label{tab:physionet_lambda}
\end{table*}

\begin{table*}[t]
\centering
\caption{Variable-specific decay rates ($\lambda$) on the MIMIC-III dataset. Higher $\lambda$ indicates faster-changing.}
\begin{tabular}{l|c|l|c|l|c|l|c}
\toprule
Variable & $\lambda$ & Variable & $\lambda$ & Variable & $\lambda$ & Variable & $\lambda$ \\
\midrule
Weight       & 6098.3438 & HR           & 11.8839  & MAP         & 138.8822 & DBP          & 4277.9697 \\
SBP          & 581.1183  & SpO2         & 2053.2156 & RR          & 11829.0942 & CRR        & 517.8212  \\
Glucose      & 17.7346   & pH           & 182.4216 & Temperature & 163.4556 & FiO2        & 18.1914   \\
GCS-EO       & 14.107    & GCS-MR       & 26.0087  & GCS-T       & 10.5168  & GCS-VR      & 15.660    \\
\bottomrule
\end{tabular}
\label{tab:mimiciii_lambda}
\end{table*}

Similar patterns are observed in P19, Physionet, and MIMIC-III. We also provided Kruskal–Wallis test results as Table~\ref{tab:kruskal_wallis_summary}. Kruskal--Wallis tests across all datasets indicate that variable-specific decay rates ($\lambda$) differ significantly (all $p \ll 0.05$). This confirms the presence of \emph{variable decay irregularity}, i.e., different clinical variables exhibit distinct temporal dynamics. For example, high-$\lambda$ variables such as Lactate and ALP change rapidly over time, whereas low-$\lambda$ variables like Creatinine and Albumin are relatively stable. These differences suggest that predictive models must account for variable-specific temporal behaviors to achieve robust clinical predictions. Notably, the MIMIC-III dataset shows the largest test statistic, indicating the strongest heterogeneity in decay rates among its variables.

\begin{table}[t]
\centering
\caption{Kruskal–Wallis test results for variable-specific decay rates ($\lambda$) across datasets.}
\begin{tabular}{l|c|c}
\toprule
Dataset      & Statistic & p-value \\
\midrule
P12          & 668.4338  & 3.87e-118 \\
P19          & 446.4491  & 5.97e-74  \\
Physionet    & 267.0967  & 5.56e-38  \\
MIMIC-III    & 697.995   & 4.97e-139 \\
\bottomrule
\end{tabular}
\label{tab:kruskal_wallis_summary}
\end{table}

This observed heterogeneity underscores the necessity of variable-specific decay modeling. Using a uniform or hand-crafted decay assumption would fail to capture these differences, effectively treating both stable indicators and highly reactive biomarkers as if they shared the same temporal patterns. By employing our designed variable decay encoding module, the proposed method can adaptively capture these irregular dynamics, enabling a more accurate representation of clinical time series and improving performance on downstream tasks.

\section{Algorithm of DBGL}
We provide the pseudo-code of DBGL here as a better illustration.

\begin{algorithm}[h]
\caption{The pseudo-code of DBGL.}
\label{alg:dbgl}

\begin{algorithmic}[1] 
\REQUIRE Input IMTS $X$, sampling timestamps $\mathbf{T}$, binary mask matrix $M$, codebook size $K$
\ENSURE Predicted labels $\hat{\mathbf{y}}$ 

\STATE Extract latent embedding $X'$ of input IMTS $X$ 
\STATE Initialize patient-variant bipartite graphs $\mathcal{G}$ from $X'$ and $M$
\FOR{each time step $t$ in $X$}
    \IF{$t>$  1}
    \STATE Conduct codebook soft-fusion as Eq.~\ref{eq:codebook} and Eq.~\ref{eq:codefuse} 
    \STATE Aggregate hidden states to patient nodes for time-step representation $h^t$
    \ENDIF
    \FOR{each variable $v$ in $G$}
        \STATE Obtain the edge embedding $E$ from $X'$ 
        \STATE Pass message over the graph and aggregation as Eq.~\ref{eq:message} and Eq.~\ref{eq:update}.
        \STATE Compute node-specific decay: $h_v^t \leftarrow h_v^{t-1} \cdot \exp(-\gamma_v \Delta t)$
        \STATE Update hidden state with current observation (edge feature) $e_{p,n}^t$ as Eq.~\ref{eq:gate}
    \ENDFOR
\ENDFOR
\STATE Obtain matched codebook representation as Eq.~\ref{eq:code_final}
\STATE Concatenate learned features for $\mathbf{z}_p$ and predict labels $\hat{y_{p}} = MLP(\mathbf{z}_p)$ as Eq.~\ref{eq:concat}.
\STATE \textbf{return} $\hat{y}_{p}$
\end{algorithmic}
\end{algorithm}

\section{Evaluated Metrics}
\textbf{AUROC.} In irregular clinical time-series classification, AUROC measures the model’s overall ability to distinguish positive from negative samples across all thresholds, reflecting its ranking or discrimination power. Even under sparse or unevenly sampled observations, AUROC provides a stable assessment of how well the model separates positive and negative cases.

\textbf{AUPRC.} In contrast, AUPRC focuses on the precision-recall trade-off, which is especially critical in clinical datasets where positive events are rare. High AUPRC indicates that the model can capture most true positive events while keeping false positives low. In irregular and imbalanced medical time-series, accurate identification of these rare but clinically important events is often more consequential than overall ranking, making AUPRC a more relevant metric than AUROC for evaluating predictive performance in such settings.

\section{Baselines}
\label{ap2}
Following \cite{luo2024knowledge}, all baseline models are implemented according to the descriptions in their original papers or the default configurations of their official code repositories. The details are as follows:

GRU-D \cite{che2018recurrent}: Based on the advanced Gated Recurrent Unit (GRU), this method incorporates two representations of missing patterns—masking and time intervals—and seamlessly integrates them into the model architecture. Code: https://github.com/Han-JD/GRU-D.

ODE-RNN \cite{rubanova2019latent}: This model uses neural ordinary differential equations (ODEs) to model hidden state dynamics and employs an RNN to update the hidden state when new observations are encountered. Code: https://github.com/YuliaRubanova/latent\_ode.

IP-Net \cite{shukla2019interpolation}: A model based on a multi-layer semi-parametric interpolation structure, which processes irregular sequences using an interpolation network followed by a predictive GRU network. Code: https://github.com/mlds-lab/interp-net.

SeFT \cite{horn2020set}: This approach uses a set function method where each observation is modeled independently before being aggregated via an attention mechanism. Code: https://github.com/BorgwardtLab/SeFT.

mTAND \cite{shukla2021multi}: A deep learning framework designed for irregular multivariate time series, which employs continuous-time embedding learning and an attention mechanism to produce fixed-length representations. Code: https://github.com/reml-lab/mTAN.

DGM$^2$-O \cite{wu2021dynamic}: A generative model that captures the evolution of latent clusters rather than independent feature representations, enabling robust modeling of sparse time series. Code: https://github.com/thuwuyinjun/DGM2.

StraTS \cite{tipirneni2022self}: A self-supervised Transformer model for sparse and irregular multivariate time series. Code: https://github.com/sindhura97/STraTS.

DuETT \cite{labach2023duett}: A dual-event time Transformer model designed for Electronic Health Records (EHRs). Code: https://github.com/layer6ai-labs/DuETT.

ViTST \cite{li2023time}: This method converts irregular multivariate time series into line graph images and adapts vision Transformer models to handle time series classification like to image classification. Code: https://github.com/Leezekun/ViTST.

Warpformer \cite{zhang2023warpformer}: A Transformer-based network that extracts multi-scale features using deformable warping modules and a dual-attention mechanism. Code: https://github.com/imJiawen/Warpformer.

MTGNN \cite{wu2020connecting}: A general graph neural network framework specifically designed for multivariate time series. Code: https://github.com/nnzhan/MTGNN

Raindrop \cite{zhang2021graph}: A graph neural network-based model that learns sensor dynamics purely from observed data. Code: https://github.com/mimsharvard/Raindrop

ISIM \cite{chen2025integrating}: A novel method to model the irregular time series data as an image and incorporate both sequence and image representations for a more generalizable joint representation. Since there was no code, we directly listed the results in the paper. 

TimeCHEAT \cite{liu2025timecheat}: A method to use the channel-dependent strategy locally and the channel-independent strategy globally for better learning of irregularly multivariate time series. Since there was no code, we directly listed the results in the paper.

KEDGN \cite{luo2024knowledge}: A method to use a pretrained language model for semantic representation extraction of each variable from the textual medical knowledge. Code:https://github.com/qianlima-lab/KEDGN.

\section{COMPUTATIONAL COSTS and bottleneck analysis}

\subsection{Comparison of computational costs.}
We analyzes of the time and space overhead on the Physionet dataset, with a batch size of 128, following \cite{luo2024knowledge}. We compare the training time per epoch (min/epoch) and the used space on GPU (MiB) as shown in Table~\ref{tab:cost}.

\begin{table}[htbp]
\centering
\caption{Training time and memory consumption per epoch of different models.}
\label{tab:efficiency}
\begin{tabular}{lccc}
\toprule
\textbf{Model} & \textbf{Time (min/epoch)} & \textbf{Space (MiB)} &  \textbf{AUPRC (\%)} \\
\midrule
ODE-RNN      & 5.06 & 2582  & 33.7 $\pm$ 4.1 \\
GRU-D        & 1.32 & 796 & 42.7 $\pm$ 7.2  \\
SeFT         & 0.07 & 684  & 29.4 $\pm$ 0.9 \\
mTAND        & 0.05 & 4658 & 52.5 $\pm$ 1.3 \\
DGM$^2$-O       & 0.06 & 684 & 50.4 $\pm$ 3.2 \\
Raindrop     & 0.17 & 4864 & 41.2 $\pm$ 3.6 \\
Warpformer   & 0.33 & 11084 & 43.5 $\pm$ 2.3 \\
KEDGN        & 0.44 & 1798 & 57.5 $\pm$ 2.5 \\
\midrule
\rowcolor[gray]{0.95} DBGL         & 0.52 & 2674 & 60.8 $\pm$ 2.2 \\
\bottomrule
\end{tabular}
\label{tab:cost}
\end{table}

As shown in the results, our proposed DBGL achieves the highest average AUPRC (60.8\% $\pm$ 2.2\%) on the Physionet dataset, significantly outperforming all baseline models and demonstrating its superior ability to distinguish between positive and negative samples. Despite its strong performance, DBGL maintains a moderate training time per epoch (0.52 min) and reasonable GPU memory usage (2674 MiB), indicating an effective trade-off between predictive accuracy and computational efficiency.

In comparison, sequential models such as ODE-RNN and GRU-D exhibit longer training times with limited predictive performance (AUPRC of 33.7\% and 42.7\%, respectively). Lightweight models like SeFT and DGM$^2$-O offer low computational overhead but achieve lower AUPRC values. Transformer-based models such as Warpformer achieve moderate performance (43.5\%) but incur substantially higher memory costs (11084 MiB), limiting their scalability. Other graph-based models, including KEDGN and mTAND, perform well but still fall short of DBGL in AUPRC. In addition, KEDGN also requires the pre-trained language model to be used in advance for variable semantic extraction, which is an additional computational overhead. Overall, these results demonstrate that DBGL provides state-of-the-art predictive performance while maintaining computational efficiency, making it highly suitable for large-scale clinical datasets and practical deployment scenarios.

\subsection{bottleneck for scaling}

The proposed patient–variable bipartite graph is constructed on a per-batch basis. Specifically, we conducted experiments on a single 24GB NVIDIA GeForce RTX 4090 GPU using the P12 dataset, exploring the impact of batch size (B), codebook size (C), sequence length (L), and number of variables (V). To vary L and V, we repeated sequences and variables accordingly to increase sequence length and variable count. We report both GPU memory consumption and training time per epoch. Our baseline configuration is B=256, V=36, L=215, C=4096, and when varying one parameter, the others are kept constant.

\begin{table*}[t]
\centering
\caption{GPU memory usage and training time per epoch under different settings (B=batch size, V=number of variables, L=sequence length, C=codebook size).}
\resizebox{\linewidth}{!}{
\begin{tabular}{c|cc|cc|cc|cc|cc|cc}
\toprule
B & Space (MiB) & Time (s/epoch) & V & Space (MiB) & Time (s/epoch) & L & Space (MiB) & Time (s/epoch) & C & Space (MiB) & Time (s/epoch) \\
\midrule
128  & 2670  & 56   & 36   & 3806   & 30   & 215  & 3806   & 30   & 512   & 2950   & 28 \\
256  & 3806  & 30   & 72   & 5258   & 35   & 430  & 6294   & 66   & 1024  & 3064   & 28 \\
512  & 6034  & 18   & 144  & 8154   & 48   & 860  & 11254  & 192  & 2048  & 3238   & 29 \\
1024 & 10776 & 12   & 288  & 14208  & 71   & 1290 & 16316  & 338  & 4096  & 3806   & 30 \\
2048 & 19700 & 9    & 432  & 20306  & 89   & 1720 & 21288  & 525  & 8172  & 4968   & 31 \\
\bottomrule
\end{tabular}
}
\label{tab:gpu_bottleneck}
\end{table*}

The results show that increasing batch size B significantly increases GPU memory usage (e.g., from 2670 MiB at B=128 to 19700 MiB at B=2048), while training time per epoch decreases due to higher GPU utilization at larger batch sizes. Increasing the number of variables V also substantially increases memory usage (from 3806 MiB at V=36 to 20306 MiB at V=432) and training time (from 30 s to 89 s per epoch). The sequence length L has the most pronounced effect on both memory and training time; for example, increasing L from 215 to 1720 raises memory from 3806 MiB to 21288 MiB and epoch time from 30 s to 525 s. In contrast, codebook size C has a relatively minor impact on memory and training time (e.g., increasing C from 512 to 8172 only increases memory from 2950 MiB to 4968 MiB, with negligible change in epoch time).

Based on these results, we analyze potential scalability bottlenecks:
\begin{itemize}
   \item GPU memory limitation is the primary constraint. Increasing batch size, number of variables, or sequence length quickly consumes GPU memory, especially in large-scale graphs or long-sequence scenarios, potentially exceeding hardware limits.
  \item Training time grows nonlinearly with sequence length; for very long sequences, per-epoch time may reach several hundred seconds, affecting experimental efficiency.
  \item In contrast, codebook size C has minimal effect on scalability, making it easier to increase without major overhead.
\end{itemize}

\subsection{Complexity Analysis}

\noindent\textbf{Time Complexity.} 
For an input of size $O(B \times T \times V \times D)$, where $B$ is the batch size, $T$ is the number of time steps, $V$ is the number of variable nodes, and $D$ is the hidden dimension, the per-step operations include: 
\begin{itemize}
    \item Codebook aggregation: $O((B+V) \cdot D \cdot N)$,
    \item GNN message passing (EdgeSAGEConv + edge update): $O(L \cdot E \cdot (D+E))$,
    \item Decay computation and gated update: $O(B \cdot V \cdot D)$,
\end{itemize}
where $L$ is the number of GNN layers, $E$ is the number of edges (smaller than $B \cdot V$ for irregular time series), and $N$ is the codebook size. Iterating over $T$ time steps gives the total time complexity:
\[
O\Big(T \cdot \big(L \cdot E \cdot (D+E) + B \cdot V \cdot D + (B+V) \cdot D \cdot N \big) \Big).
\]
While our method introduces additional overhead via codebook aggregation and EdgeSAGEConv, both GNN message passing and node decay updates are linear in $BVD$ and can be efficiently parallelized on GPUs.

\noindent\textbf{Space Complexity.} 
The memory cost for storing inputs, hidden states, codebook, and edge attributes is
\[
O(B \cdot T \cdot V \cdot D + E \cdot D + N \cdot D).
\]

\subsection{Runtime Comparison}

To quantify the computational overhead, we compare our model (DBGL) with the sequence-based baseline KEDGN using the same GPU (a NVIDIA A800 80GB PCIe). The measured testing time (in seconds) is summarized in Table~\ref{tab:runtime_test} and Table~\ref{tab:runtime_train}.

\begin{table}[h]
\centering
\caption{Inference time (seconds) of DBGL and KEDGN on four datasets.}
\begin{tabular}{l|cccc}
\toprule
Model & P19 & Physionet & MIMIC-III & P12 \\
\midrule
KEDGN & 0.11 & 0.14 & 0.61 & 0.29 \\
DBGL  & 0.18 & 0.16 & 0.84 & 0.50 \\
\bottomrule
\end{tabular}
\label{tab:runtime_test}
\end{table}

\begin{table}[h]
\centering
\caption{Training time per batch (seconds) of DBGL and KEDGN on four datasets.}
\begin{tabular}{l|cccc}
\toprule
Model & P19 & Physionet & MIMIC-III & P12 \\
\midrule
KEDGN & 1.76 & 0.93 & 5.92 & 1.61 \\
DBGL  & 2.03 & 0.81 & 4.86 & 1.92 \\
\bottomrule
\end{tabular}
\label{tab:runtime_train}
\end{table}

Several observations can be drawn from the results:
\begin{itemize}
    \item The inference time of DBGL is highly comparable to that of the sequence-based KEDGN. On P19 and P12, DBGL is slightly slower (+0.27s and +0.31s), reflecting the lightweight online graph construction. On Physionet and MIMIC-III, DBGL is even faster ($-0.12$~s and $-1.06$~s), showing that dynamic graph generation does not introduce significant computational burden.
    \item The dynamic graph building process adds only minor overhead because graph nodes correspond directly to observed variables at each step, and the graph size remains small (patient node + variable nodes), resulting in linear complexity with respect to observed variables.
    \item Across all datasets, the runtime difference between DBGL and sequence baselines is within 0.3--1.0 seconds, demonstrating that the proposed graph mechanism is both practical and efficient for real-world healthcare scenarios.
\end{itemize}

In summary, although our patient--variable graphs are dynamically generated at inference time, this process introduces minimal overhead, and the overall inference speed remains competitive and often even faster than sequence-based baselines.

\section{More experiments}

\subsection{Clinical deployment}

To better verify the clinical deployment feasibility of our proposed DBGL, we also conduct a comparative analysis on Expected Calibration Error (ECE) and Brier Score (BS) with KEDGN, evaluating the models' predictive uncertainty calibration and overall probabilistic accuracy. The comparative results across four clinical benchmarks are presented in Table~\ref{tab:calibration_results}, formatted as ECE(\%) / Brier Score(\%).

\begin{table}[htbp]
\centering
\caption{ECE (\%) and Brier Score (\%) comparison between DBGL and KEDGN across clinical datasets. Results are presented as mean $\pm$ standard deviation.}
\label{tab:calibration_results}
\begin{tabular}{lcccc}
\toprule
\textbf{Model} & \textbf{P19} & \textbf{PhysioNet} & \textbf{MIMIC-III} & \textbf{P12} \\
\midrule
KEDGN & 10.9 $\pm$ 1.7 / 21.3 $\pm$ 3.0 & 15.8 $\pm$ 1.2 / 22.5 $\pm$ 2.4 & 15.8 $\pm$ 2.1 / 21.5 $\pm$ 4.2 & 15.0 $\pm$ 1.1 / 19.8 $\pm$ 1.7 \\
DBGL  & \textbf{9.2 $\pm$ 1.6} / \textbf{18.2 $\pm$ 3.2} & \textbf{14.5 $\pm$ 1.5} / \textbf{21.1 $\pm$ 2.1} & \textbf{14.3 $\pm$ 1.0} / \textbf{20.0 $\pm$ 2.1} & \textbf{14.2 $\pm$ 1.4} / \textbf{18.9 $\pm$ 2.6} \\
\bottomrule
\end{tabular}
\end{table}

Our calibration analysis demonstrates DBGL's superior probabilistic reliability over KEDGN across all clinical benchmarks. DBGL achieves consistently lower Expected Calibration Error (ECE) and Brier Score values, with the most significant improvements observed on the P19 (3.1 Brier Score reduction) and MIMIC-III (1.5 ECE reduction) datasets. These results indicate that DBGL provides better-calibrated uncertainty estimates—its predicted confidence levels more accurately reflect true accuracy probabilities—while also delivering superior overall probabilistic forecasting performance. The enhanced calibration is particularly valuable for clinical deployment, as it ensures that risk predictions are both accurate and reliably quantified, enabling clinicians to make more informed decisions with greater trust in the model's confidence estimates.

\subsection{Different Group Patient Analysis}

We conducted extensive experiments across four datasets: P12, P19, Physionet, and MIMIC-III, each with distinct patient distributions, observation lengths, variable sets, and clinical characteristics. To examine the effect of patient cohort composition, we randomly split the training set of each dataset into two non-overlapping subsets (50\% of patients each), trained separate models on each subset, and evaluated their performance. Results are shown in Table~\ref{tab:subset_performance_formatted}.

\begin{table*}[t]
\centering
\caption{Performance on different training subsets across four clinical datasets (AUROC \& AUPRC, \%).}
\resizebox{\linewidth}{!}{
\begin{tabular}{l|cc|cc|cc|cc}
\toprule
Subset & \multicolumn{2}{c}{P19} & \multicolumn{2}{c}{Physionet} & \multicolumn{2}{c}{MIMIC-III} & \multicolumn{2}{c}{P12} \\
\cmidrule(lr){2-3} \cmidrule(lr){4-5} \cmidrule(lr){6-7} \cmidrule(lr){8-9}
& AUROC & AUPRC & AUROC & AUPRC & AUROC & AUPRC & AUROC & AUPRC \\
\midrule
Subset 1 & 90.8 $\pm$ 0.3 & 61.0 $\pm$ 1.6 & 87.0 $\pm$ 1.5 & 55.1 $\pm$ 4.5 & 83.7 $\pm$ 0.5 & 46.1 $\pm$ 1.2 & 85.9 $\pm$ 0.6 & 50.8 $\pm$ 1.0 \\
Subset 2 & 91.0 $\pm$ 0.2 & 61.6 $\pm$ 1.2 & 86.5 $\pm$ 2.6 & 54.6 $\pm$ 3.3 & 84.0 $\pm$ 0.3 & 45.8 $\pm$ 1.3 & 85.7 $\pm$ 0.7 & 50.8 $\pm$ 1.9 \\
\rowcolor[gray]{0.95} All      & 93.3 $\pm$ 0.5 & 66.3 $\pm$ 1.4 & 89.1 $\pm$ 0.3 & 60.8 $\pm$ 2.2 & 85.2 $\pm$ 0.4 & 50.2 $\pm$ 1.0 & 88.1 $\pm$ 0.4 & 56.3 $\pm$ 1.0 \\
\bottomrule
\end{tabular}
}
\label{tab:subset_performance_formatted}
\end{table*}

We found that performance gaps between models trained on different subsets are minimal, demonstrating strong generalization across patient distributions. Even with only 50\% of the training data, our model still outperforms nearly all baselines and is only slightly below KEDGN, supporting the robustness and effectiveness of our design.

Since P12, P19, and Physionet include ICU-type labels, we further evaluated model performance across different patient groups.

\begin{table}[h]
\centering
\caption{P19 dataset: performance across ICU types.}
\scalebox{0.8}{
\begin{tabular}{lcc}
\toprule
ICU Type & Samples & AUROC \\
\midrule
Medical ICU & 3872 & 93.2$\pm$0.5 \\
SIC Surgical ICU & 9 & 98.6$\pm$2.9 \\
\bottomrule
\end{tabular}
}
\end{table}

\begin{table}[h]
\centering
\caption{P12 dataset: performance across ICU types.}
\scalebox{0.8}{
\begin{tabular}{lcccc}
\toprule
ICU Type & Coronary Care Unit & Cardiac Surgery Recovery Unit & Medical ICU & SIC Surgical ICU \\
\midrule
Samples & 219 & 231 & 419 & 339 \\
AUROC & 81.9$\pm$1.5 & 83.4$\pm$1.9 & 85.0$\pm$0.5 & 90.4$\pm$0.3 \\
\bottomrule
\end{tabular}}
\end{table}

\begin{table}[h]
\centering
\caption{Physionet dataset: performance across ICU types.}
\scalebox{0.8}{
\begin{tabular}{lcccc}
\toprule
ICU Type & Coronary Care Unit & Cardiac Surgery Recovery Unit & Medical ICU & SIC Surgical ICU \\
\midrule
Samples & 64 & 91 & 153 & 91 \\
AUROC & 90.3$\pm$2.4 & 99.4$\pm$0.6 & 84.8$\pm$0.5 & 85.4$\pm$1.2 \\
\bottomrule
\end{tabular}
}
\end{table}

Across different ICU types and datasets, the model maintains strong predictive performance, demonstrating good generalization across diverse patient populations. Extremely high AUROC values often occur in ICU types with small sample sizes or highly homogeneous patient characteristics (e.g., cardiac surgery recovery patients), where clinical patterns are more consistent. While performance is stable, small-sample groups may introduce higher variance.

Overall, these results show that our model is robust, generalizes well across heterogeneous patient cohorts, and performs particularly strongly in clinically consistent populations.

\subsection{Experiments about LEAVE-VARIABLES-OUT.}
\label{ap:31}

We provide more results about Leave-Variables-Out experiments on MIMIC-III, P19, and Physionet datasets here. The experimental results are shown in Table~\ref{table:res3}.

\begin{table*}[t]
\centering
\caption{Performance comparison (AUROC \& AUPRC, \%) under different discarding ratios on three clinical datasets. The best results are highlighted in \textbf{bold} and the second-best results are in \underline{underlined}.}
\resizebox{\linewidth}{!}{
\begin{tabular}{l|l|cc|cc|cc|cc|cc}
\toprule
& \multirow{2}{*}{Method} 
& \multicolumn{2}{c}{10\%} & \multicolumn{2}{c}{20\%} & \multicolumn{2}{c}{30\%} & \multicolumn{2}{c}{40\%} & \multicolumn{2}{c}{50\%} \\
\cmidrule(lr){3-4}\cmidrule(lr){5-6}\cmidrule(lr){7-8}\cmidrule(lr){9-10}\cmidrule(lr){11-12}
& & AUROC & AUPRC & AUROC & AUPRC & AUROC & AUPRC & AUROC & AUPRC & AUROC & AUPRC \\
\midrule
\multirow{9}{*}{MIMIC-III}
& GRU-D      & 81.0 $\pm$ 0.6 & 42.1 $\pm$ 0.8 & 80.3 $\pm$ 0.9 & 41.7 $\pm$ 1.0 & 79.2 $\pm$ 1.8 & 41.0 $\pm$ 1.4 & 78.5 $\pm$ 2.1 & 40.4 $\pm$ 1.6 & 77.9 $\pm$ 2.2 & 39.9 $\pm$ 1.8 \\
& mTAND      & 81.2 $\pm$ 0.2 & 42.1 $\pm$ 0.8 & 80.4 $\pm$ 1.1 & 41.9 $\pm$ 1.2 & 79.7 $\pm$ 1.4 & 41.0 $\pm$ 1.7 & 79.3 $\pm$ 1.4 & 40.4 $\pm$ 2.0 & 78.8 $\pm$ 1.6 & 39.8 $\pm$ 2.3 \\
& $\text{DGM}^{2}$-O     & 78.8 $\pm$ 0.5 & 34.2 $\pm$ 0.9 & 78.3 $\pm$ 0.8 & 33.9 $\pm$ 1.1 & 77.6 $\pm$ 1.2 & 33.4 $\pm$ 1.2 & 77.3 $\pm$ 1.3 & 33.1 $\pm$ 1.2 & 76.8 $\pm$ 1.5 & 32.6 $\pm$ 1.4 \\
& MTGNN      & 78.8 $\pm$ 1.1 & 34.5 $\pm$ 1.4 & 78.0 $\pm$ 1.6 & 34.0 $\pm$ 1.3 & 77.1 $\pm$ 2.2 & 33.5 $\pm$ 1.5 & 76.3 $\pm$ 2.5 & 32.8 $\pm$ 1.9 & 75.6 $\pm$ 3.2 & 32.2 $\pm$ 2.4 \\
& Raindrop   & 78.2 $\pm$ 1.1 & 33.7 $\pm$ 0.9 & 77.5 $\pm$ 1.3 & 33.5 $\pm$ 0.9 & 76.4 $\pm$ 2.1 & 32.8 $\pm$ 1.4 & 76.0 $\pm$ 2.0 & 32.5 $\pm$ 1.4 & 75.7 $\pm$ 2.0 & 32.3 $\pm$ 1.4 \\
& DuETT      & 78.0 $\pm$ 0.5 & 34.0 $\pm$ 0.9 & 77.2 $\pm$ 1.0 & 33.7 $\pm$ 0.8 & 76.6 $\pm$ 1.2 & 33.3 $\pm$ 1.0 & 76.4 $\pm$ 1.2 & 33.0 $\pm$ 1.0 & 76.1 $\pm$ 1.3 & 32.6 $\pm$ 1.3 \\
& Warpformer & 82.5 $\pm$ 0.5 & 43.1 $\pm$ 0.8 & 81.7 $\pm$ 0.9 & 42.5 $\pm$ 1.2 & 81.2 $\pm$ 1.1 & 42.1 $\pm$ 1.2 & 80.6 $\pm$ 1.5 & 41.8 $\pm$ 1.3 & 80.0 $\pm$ 1.9 & 41.3 $\pm$ 1.6 \\
& KEDGN      & \underline{83.0 $\pm$ 0.7} & \underline{44.8 $\pm$ 2.0} & \underline{82.3 $\pm$ 1.1} & \underline{44.4 $\pm$ 1.9} & \underline{81.3 $\pm$ 1.9} & \underline{43.6 $\pm$ 2.2} & \underline{80.6 $\pm$ 2.1} & \underline{43.0 $\pm$ 2.4} & \underline{80.0 $\pm$ 2.3} & \underline{42.4 $\pm$ 2.5} \\
\rowcolor[gray]{0.95} & DBGL       &      \textbf{84.3 $\pm$ 0.3}        &   \textbf{48.2 $\pm$ 1.3}           & \textbf{83.4 $\pm$ 0.6}            &      \textbf{45.7 $\pm$ 1.2 }        &    \textbf{83.4 $\pm$ 0.7}         &  \textbf{45.2 $\pm$ 0.7}           &       \textbf{82.3 $\pm$ 0.8}      &          \textbf{43.9 $\pm$ 1.5 }   &  \textbf{81.1 $\pm$ 0.9}        &    \textbf{42.9 $\pm$ 1.0}          \\
\midrule
\multirow{9}{*}{P19}
& GRU-D & 88.5 $\pm$ 2.3 & 54.6 $\pm$ 3.7 & 88.8 $\pm$ 2.1 & 54.2 $\pm$ 3.4 & 88.0 $\pm$ 2.5 & 50.4 $\pm$ 7.5 & 87.5 $\pm$ 2.8 & 49.6 $\pm$ 6.9 & 86.4 $\pm$ 3.5 & 47.2 $\pm$ 8.6 \\
& mTAND & 79.6 $\pm$ 1.8 & 28.6 $\pm$ 1.9 & 79.2 $\pm$ 1.9 & 28.1 $\pm$ 2.1 & 78.0 $\pm$ 2.4 & 26.9 $\pm$ 2.9 & 77.2 $\pm$ 2.7 & 26.3 $\pm$ 2.9 & 76.2 $\pm$ 3.2 & 24.3 $\pm$ 4.8 \\
& $\text{DGM}^{2}$-O & 87.4 $\pm$ 0.6 & 53.4 $\pm$ 1.5 & 87.3 $\pm$ 0.8 & 53.2 $\pm$ 1.7 & 86.6 $\pm$ 1.6 & 49.9 $\pm$ 5.1 & 85.8 $\pm$ 1.9 & 47.7 $\pm$ 5.9 & 85.2 $\pm$ 2.2 & 45.7 $\pm$ 6.7 \\
& MTGNN & 84.5 $\pm$ 1.4 & 48.9 $\pm$ 2.3 & 84.8 $\pm$ 1.7 & 49.8 $\pm$ 3.1 & 84.0 $\pm$ 1.9 & 47.2 $\pm$ 4.8 & 83.3 $\pm$ 2.2 & 45.5 $\pm$ 5.5 & 82.5 $\pm$ 2.9 & 42.7 $\pm$ 9.2 \\
& Raindrop & 88.2 $\pm$ 1.5 & 59.7 $\pm$ 1.5 & 88.1 $\pm$ 1.3 & 59.8 $\pm$ 1.4 & 87.8 $\pm$ 1.2 & 59.1 $\pm$ 1.7 & 87.6 $\pm$ 1.1 & 58.5 $\pm$ 1.9 & 87.1 $\pm$ 1.5 & 87.1 $\pm$ 1.5 \\
& DuETT & 85.2 $\pm$ 1.0 & 53.7 $\pm$ 1.0 & 84.8 $\pm$ 1.1 & 53.9 $\pm$ 0.8 & 84.7 $\pm$ 1.0 & 53.3 $\pm$ 1.6 & 84.3 $\pm$ 1.4 & 52.7 $\pm$ 2.1 & 84.4 $\pm$ 1.3 & 52.5 $\pm$ 2.0 \\
& Warpformer & 91.3 $\pm$ 0.8 & 55.2 $\pm$ 5.6 & \underline{91.3 $\pm$ 0.8} & 55.1 $\pm$ 5.6 & \underline{91.4 $\pm$ 0.8} & 56.0 $\pm$ 4.8 & 91.5 $\pm$ 0.7 & 56.4 $\pm$ 4.3 & 91.2 $\pm$ 0.8 & 56.2 $\pm$ 3.9 \\
& KEDGN & \underline{91.3 $\pm$ 0.3} & \underline{59.9 $\pm$ 0.7} & 91.2 $\pm$ 0.5 & \underline{59.6 $\pm$ 0.9} & 90.9 $\pm$ 0.9 & \underline{59.1 $\pm$ 1.1} & 90.7 $\pm$ 1.0 & 58.9 $\pm$ 1.2 & 90.1 $\pm$ 1.6 & 58.1 $\pm$ 2.0 \\
\rowcolor[gray]{0.95} & DBGL & \textbf{92.8 $\pm$ 0.8} & \textbf{64.4 $\pm$ 1.2} & \textbf{91.5 $\pm$ 0.8} & \textbf{61.8 $\pm$ 1.4} & \textbf{91.4 $\pm$ 0.5} & \textbf{60.8 $\pm$ 2.6} & \textbf{91.2 $\pm$ 0.4} & \textbf{59.2 $\pm$ 1.8} & \textbf{90.4 $\pm$ 0.3} & \textbf{58.3 $\pm$ 1.8} \\
\midrule
\multirow{9}{*}{Physionet}
& GRU-D & 70.0 $\pm$ 3.0 & 32.1 $\pm$ 4.1 & 69.5 $\pm$ 2.6 & 31.1 $\pm$ 3.6 & 69.2 $\pm$ 3.0 & 31.0 $\pm$ 4.4 & 68.3 $\pm$ 3.6 & 30.1 $\pm$ 5.3 & 68.1 $\pm$ 3.7 & 29.8 $\pm$ 5.3 \\
& mTAND & 80.5 $\pm$ 2.1 & 42.8 $\pm$ 4.0 & 78.2 $\pm$ 3.4 & 40.5 $\pm$ 4.7 & 76.3 $\pm$ 4.0 & 37.7 $\pm$ 5.7 & 75.6 $\pm$ 3.9 & 36.6 $\pm$ 5.6 & 75.1 $\pm$ 3.9 & 36.1 $\pm$ 5.1 \\
& $\text{DGM}^{2}$-O & 80.2 $\pm$ 0.9 & 38.6 $\pm$ 2.8 & 80.4 $\pm$ 0.9 & 38.3 $\pm$ 2.8 & 79.3 $\pm$ 1.9 & 37.1 $\pm$ 3.4 & 77.5 $\pm$ 3.7 & 35.4 $\pm$ 4.4 & 75.6 $\pm$ 5.0 & 34.0 $\pm$ 4.8 \\
& MTGNN & 68.9 $\pm$ 4.1 & 25.8 $\pm$ 4.8 & 69.3 $\pm$ 4.3 & 26.6 $\pm$ 4.5 & 69.0 $\pm$ 4.8 & 26.3 $\pm$ 5.2 & 68.3 $\pm$ 5.2 & 25.4 $\pm$ 4.8 & 67.2 $\pm$ 5.4 & 24.4 $\pm$ 4.8 \\
& Raindrop & 76.5 $\pm$ 1.2 & 33.4 $\pm$ 2.2 & 76.5 $\pm$ 1.3 & 32.3 $\pm$ 2.3 & 75.6 $\pm$ 2.0 & 30.8 $\pm$ 3.2 & 74.7 $\pm$ 2.6 & 29.7 $\pm$ 3.5 & 73.6 $\pm$ 3.2 & 28.8 $\pm$ 3.9 \\
& DuETT & 78.2 $\pm$ 2.8 & 39.9 $\pm$ 3.5 & 78.3 $\pm$ 3.0 & 39.9 $\pm$ 3.7 & 76.7 $\pm$ 3.7 & 37.9 $\pm$ 4.5 & 75.9 $\pm$ 3.8 & 37.0 $\pm$ 4.6 & 74.9 $\pm$ 4.3 & 35.9 $\pm$ 5.0 \\
& Warpformer & 78.2 $\pm$ 1.0 & 33.3 $\pm$ 2.1 & 77.7 $\pm$ 1.6 & 33.6 $\pm$ 1.8 & 75.8 $\pm$ 3.4 & 31.8 $\pm$ 3.0 & 73.8 $\pm$ 4.6 & 30.2 $\pm$ 4.1 & 72.7 $\pm$ 4.9 & 29.2 $\pm$ 4.2 \\
& KEDGN & 83.8 $\pm$ 1.0 & 49.4 $\pm$ 2.5 & 82.9 $\pm$ 2.5 & 48.0 $\pm$ 5.3 & 81.7 $\pm$ 2.8 & 46.4 $\pm$ 5.5 & 81.4 $\pm$ 2.5 & 45.8 $\pm$ 5.2 & 81.1 $\pm$ 2.4 & 45.2 $\pm$ 5.3 \\
\rowcolor[gray]{0.95} & DBGL & \textbf{88.6 $\pm$ 0.9} & \textbf{57.8 $\pm$ 1.7} & \textbf{86.6 $\pm$ 1.5} & \textbf{53.6 $\pm$ 3.5} & \textbf{84.0 $\pm$ 2.2} & \textbf{50.1 $\pm$ 2.0} & \textbf{82.4 $\pm$ 1.9} & \textbf{48.1 $\pm$ 2.4} & \textbf{82.8 $\pm$ 1.5} & \textbf{46.7 $\pm$ 2.9} \\
\bottomrule
\end{tabular}
}
\label{table:res3}
\end{table*}

As shown in Table~\ref{table:res3}, DBGL consistently outperforms all baselines across different discarding ratios and datasets. On MIMIC-III, even when 50\% of the variables are removed, DBGL maintains an AUROC of 81.1\% and an AUPRC of 42.9\%, significantly higher than the second-best method KEDGN (AUROC 80.0\%, AUPRC 42.4\%). Similar trends are observed on P19, where DBGL achieves AUROC/AUPRC scores of 90.4\%/58.3\% under 50\% variable removal, outperforming Warpformer and KEDGN. On Physionet, DBGL also demonstrates superior robustness, with AUROC/AUPRC remaining at 82.8\%/46.7\% under the most extreme 50\% variable drop. These results highlight the effectiveness of DBGL in leveraging patient-variant bipartite graphs and node-specific decay mechanisms to capture temporal dynamics, even under significant feature missingness. The consistent performance gain indicates that DBGL can robustly model irregularly sampled clinical time series and is less sensitive to missing or discarded variables compared to existing state-of-the-art methods.

\subsection{Effect of the number of EdgeSAGE layers.}
\begin{figure}[htbp]
  \centering
\includegraphics[width=0.5\textwidth]{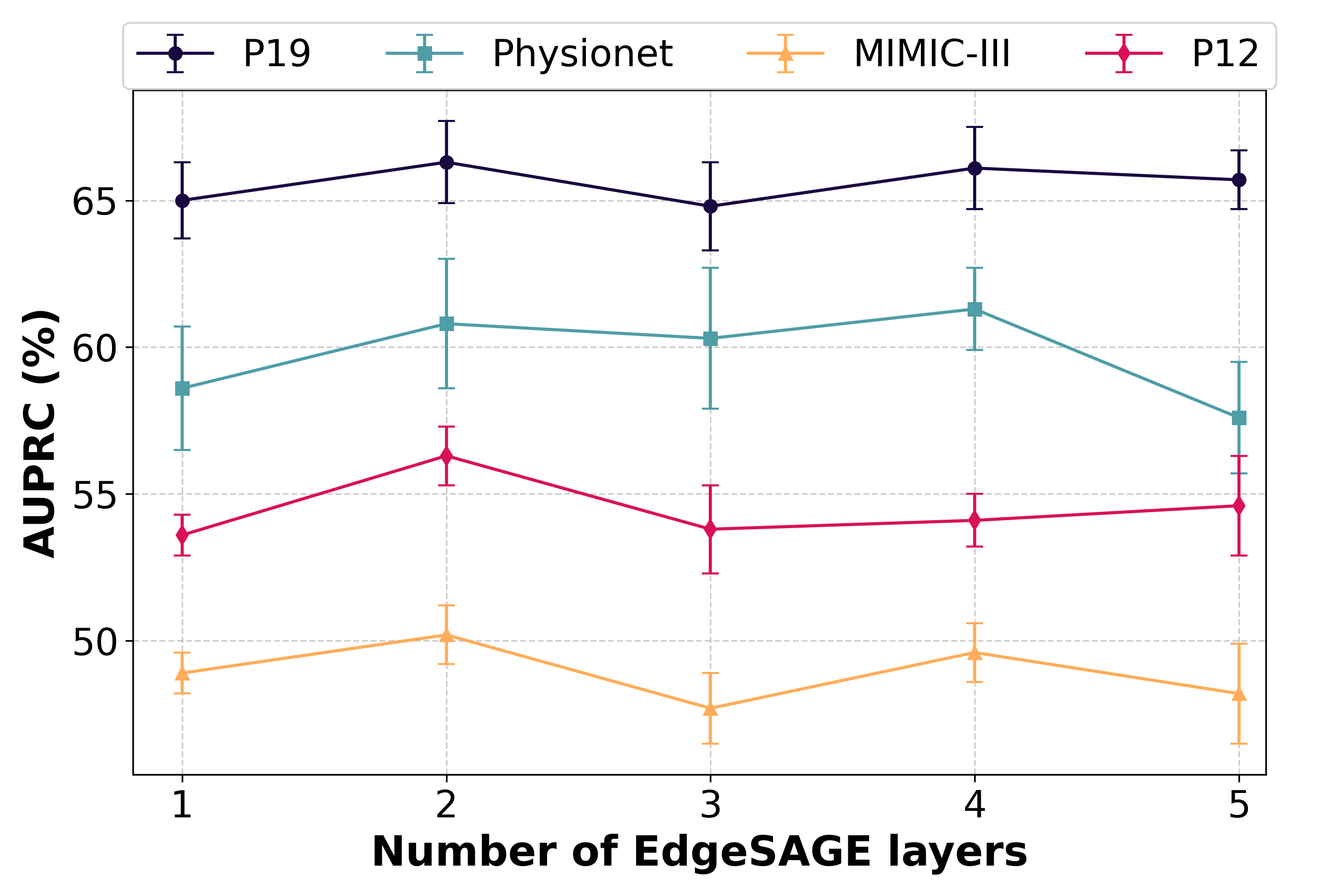}
  \caption{Effect of the number of EdgeSAGE layers.}
  \label{fig:edgesage_layers}
\end{figure}

To investigate the impact of the number of EdgeSAGE layers in DBGL, we vary the number of layers from 1 to 5 in increments of 1. We also analyze the influence of graph convolution depth, as increasing the number of GCN layers allows the model to capture higher-order dependencies between patient and variable nodes, but may also introduce over-smoothing if too deep, and bring more computation cost. The experimental results are shown in Figure~\ref{fig:edgesage_layers}.

Figure~\ref{fig:edgesage_layers} shows the AUPRC of DBGL as the number of EdgeSAGE layers varies from 1 to 5. Figure~\ref{fig:edgesage_layers} shows the AUPRC of DBGL as the number of EdgeSAGE layers varies from 1 to 5. Increasing the depth from 1 to 2 consistently improves AUPRC across all datasets (e.g., P19: 65.0\% → 66.3\%), indicating that capturing higher-order patient-variable dependencies benefits positive-sample discrimination. Beyond 2 layers, gains plateau or fluctuate slightly, suggesting diminishing returns and potential over-smoothing. Based on these results, we select a 2-layer EdgeSAGE model as the final configuration. Beyond 2 layers, gains plateau or fluctuate slightly, suggesting diminishing returns and potential over-smoothing, and also bring a larger computational cost. Based on these results, we select a 2-layer EdgeSAGE model as the final configuration.

\subsection{Effect of the size of codebook.}
To investigate the impact of codebook size on model performance, we vary the number of discrete codewords in the codebook and evaluate the resulting predictive accuracy and representation quality. A larger codebook allows the model to capture finer-grained variations in the input space, potentially improving expressiveness, but it may also increase computational cost and the risk of overfitting. Conversely, a smaller codebook enforces stronger quantization, which can act as a regularizer but may lead to information loss. The size of the codebook is set from 512 to 8192 and measured on all datasets.

\begin{figure}[htbp]
  \centering
\includegraphics[width=0.5\textwidth]{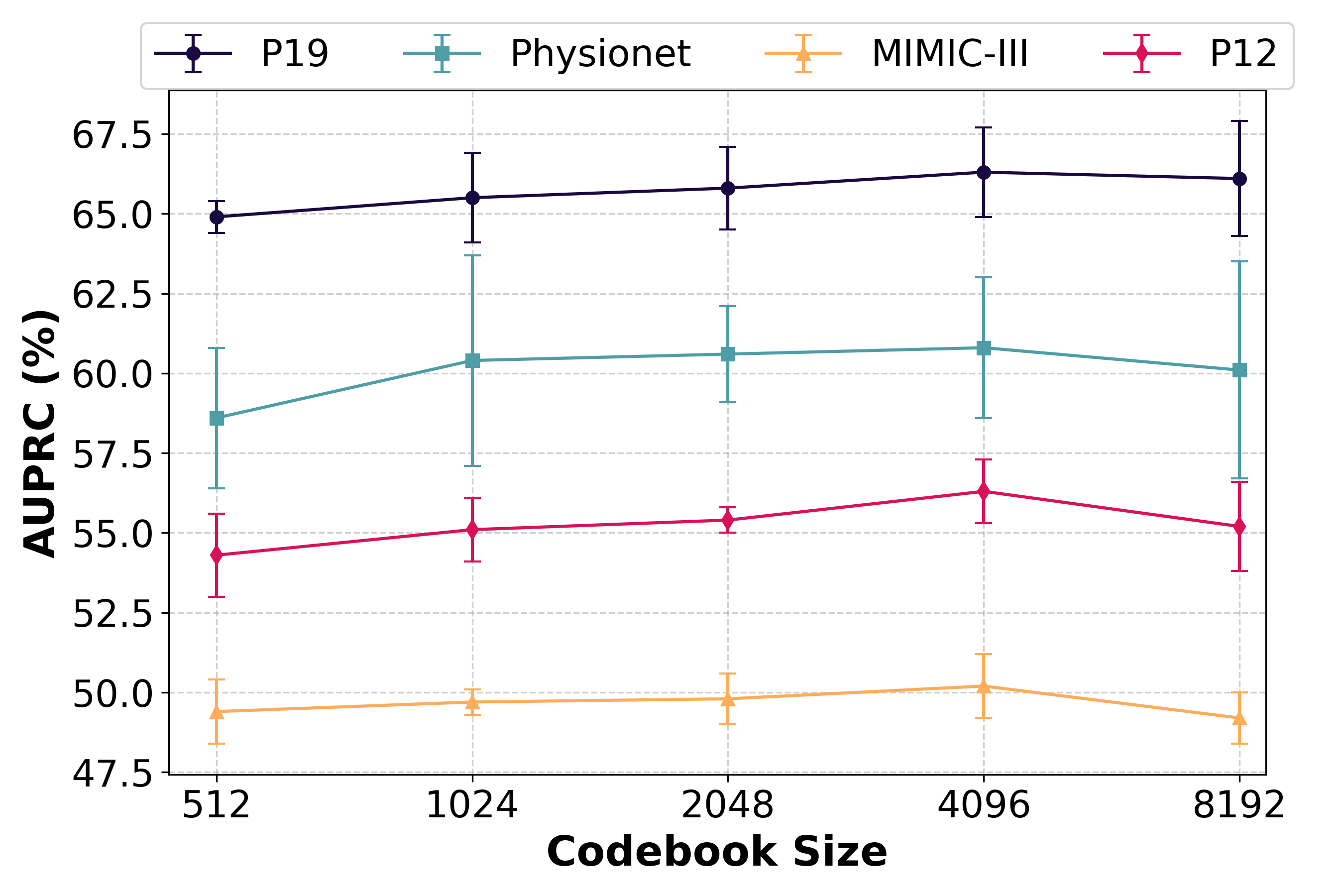}
  \caption{Comparison of Positive-class Predicted Probabilities.}
  \label{fig:cb}
\end{figure}

As shown in Figure~\ref{fig:cb}, increasing the codebook size generally improves performance initially, with the model achieving the highest AUROC and AUPRC on most datasets around 2048–4096 codewords. For instance, on P19, AUROC increases from 92.9\% to 93.3\% and AUPRC from 64.9\% to 66.3\% as the codebook grows from 512 to 4096. Similar trends are observed on P12 and MIMIC-III, while Physionet shows a slight drop beyond 2048, suggesting diminishing returns with overly large codebooks. These results indicate a trade-off between expressiveness and overfitting: a larger codebook enables finer-grained representation of input features, improving predictive power, but excessively large codebooks can introduce noise and marginally degrade performance.

To further examine codebook usage, we compute the \emph{soft utilization rate}, defined as the fraction of code vectors receiving above-uniform activation.

Given the soft assignment weights $w \in \mathbb{R}^{N \times K}$, the utilization rate is defined as:
\begin{equation}
\bar{w}_k = \frac{1}{N} \sum_{i=1}^{N} w_{i,k}, \qquad
\text{Utilization} = \frac{1}{K} \sum_{k=1}^{K} \mathbb{1} \left( \bar{w}_k > \frac{1}{K} \right),
\end{equation}
where $N$ is the number of samples and $K$ is the number of codebook entries. This metric quantifies how many codebook entries are used more frequently than what a uniform distribution would predict.

\begin{table*}[t]
\centering
\caption{Soft codebook utilization rates across different datasets.}
\begin{tabular}{lcccc}
\toprule
Dataset & P19 & Physionet & MIMIC-III & P12 \\
\midrule
Utilization Rate & 0.4973 & 0.4977 & 0.4982 & 0.4978 \\
\bottomrule
\end{tabular}
\label{tab:codebook_utilization}
\end{table*}

There are some key observations from the results:
\begin{itemize}
    \item \textbf{No collapse occurs}: utilization remains $\sim 0.498$ for all datasets, far from zero.
    \item \textbf{Balanced engagement}: code vectors are activated evenly across the codebook, showing stable and effective use.
    \item \textbf{Consistent usage across datasets}: despite differing dynamics, the codebook contributes meaningfully, particularly in homogeneous datasets where similarity-guided aggregation is more beneficial.
\end{itemize}

These results confirm that the codebook is not merely a design embellishment but a robust mechanism that facilitates selective cross-sample communication, enhances learning of variable dependencies, and improves predictive performance when similarity structure is present.

\section{Visualization}
\begin{figure}[t]
    \centering
    \begin{subfigure}{0.32\linewidth}
        \centering
        \includegraphics[width=\linewidth]{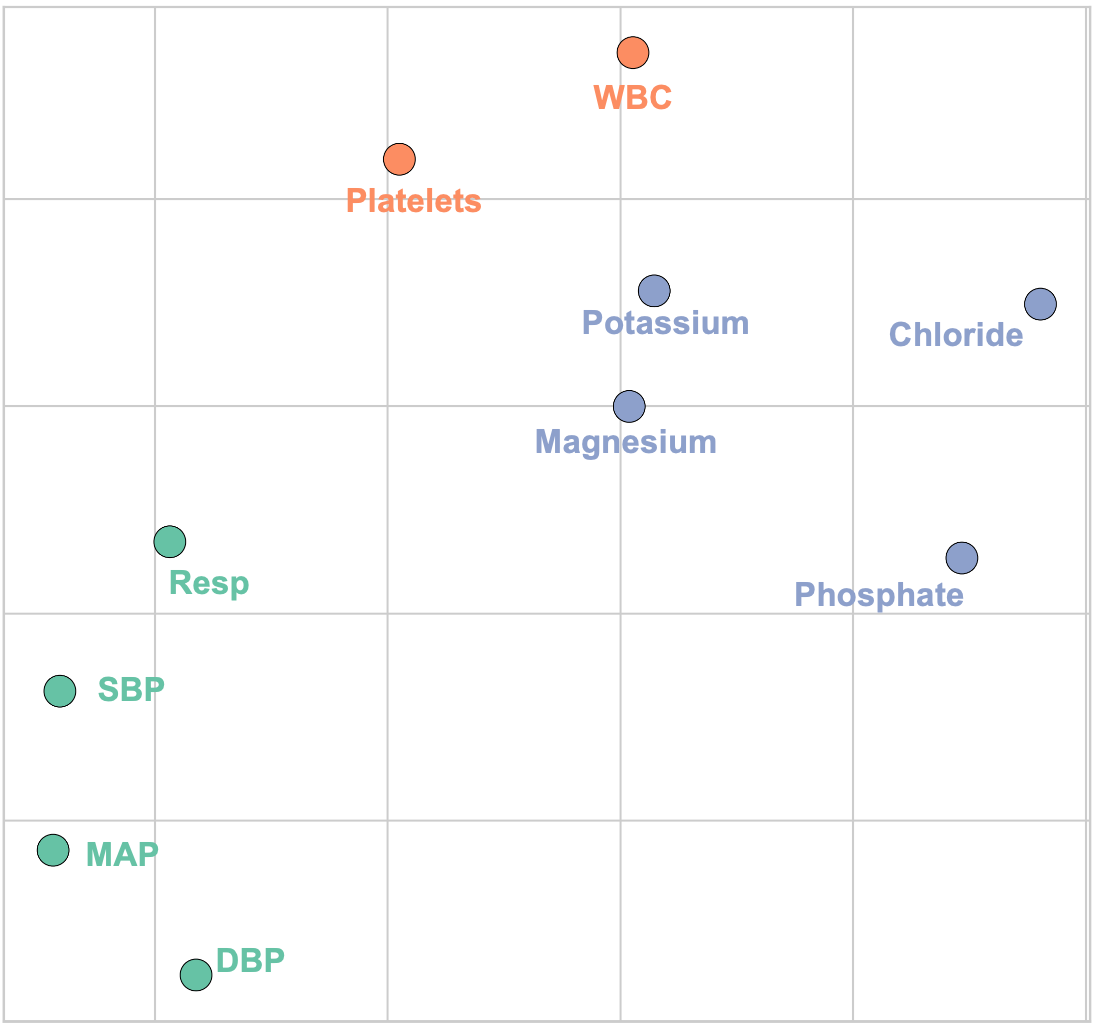}
        \caption{P19 Dataset}
        \label{fig:sub1}
    \end{subfigure}
    \hfill
    \begin{subfigure}{0.32\linewidth}
        \centering
        \includegraphics[width=\linewidth]{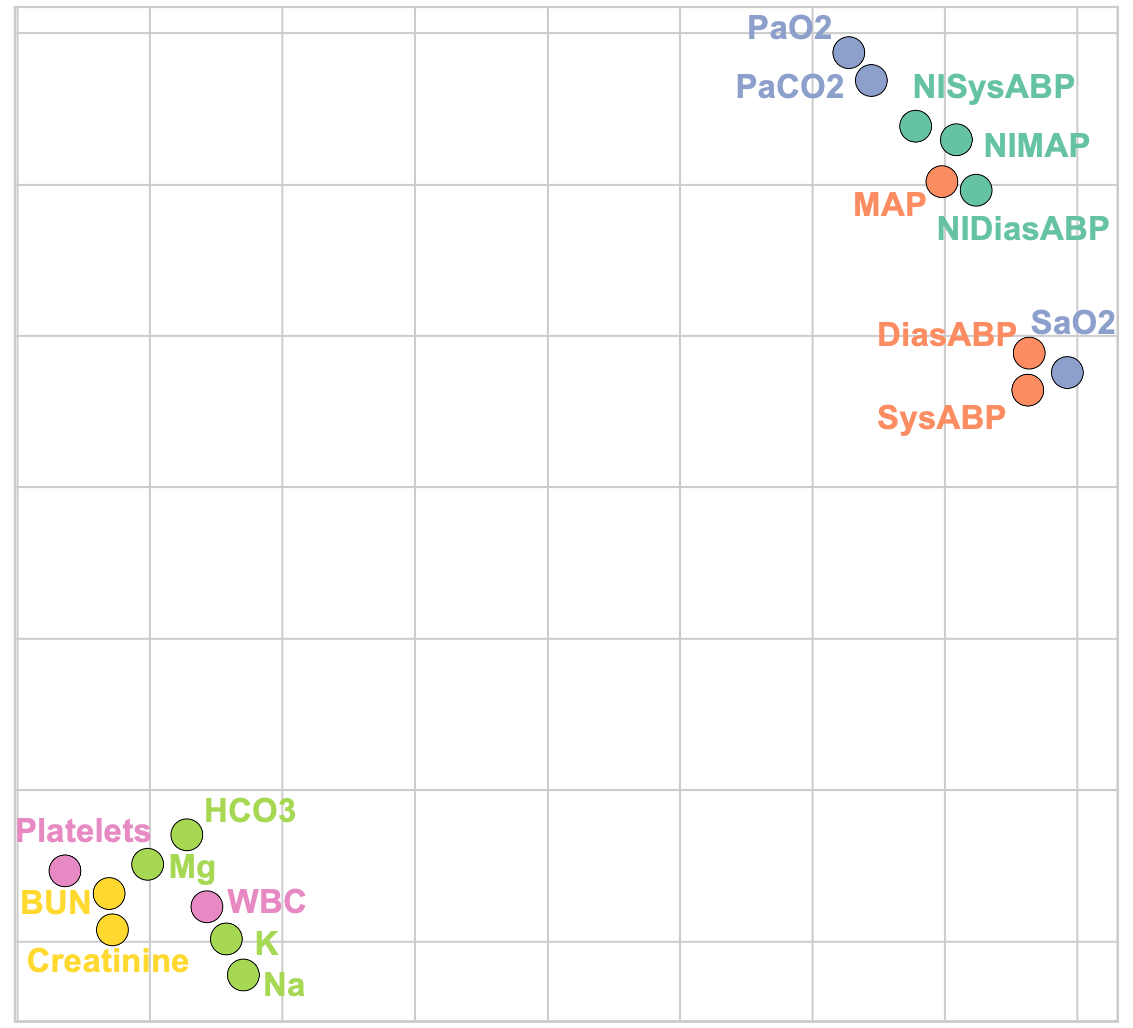}
        \caption{P12 Dataset}
        \label{fig:sub2}
    \end{subfigure}
    \hfill
    \begin{subfigure}{0.32\linewidth}
        \centering
        \includegraphics[width=\linewidth]{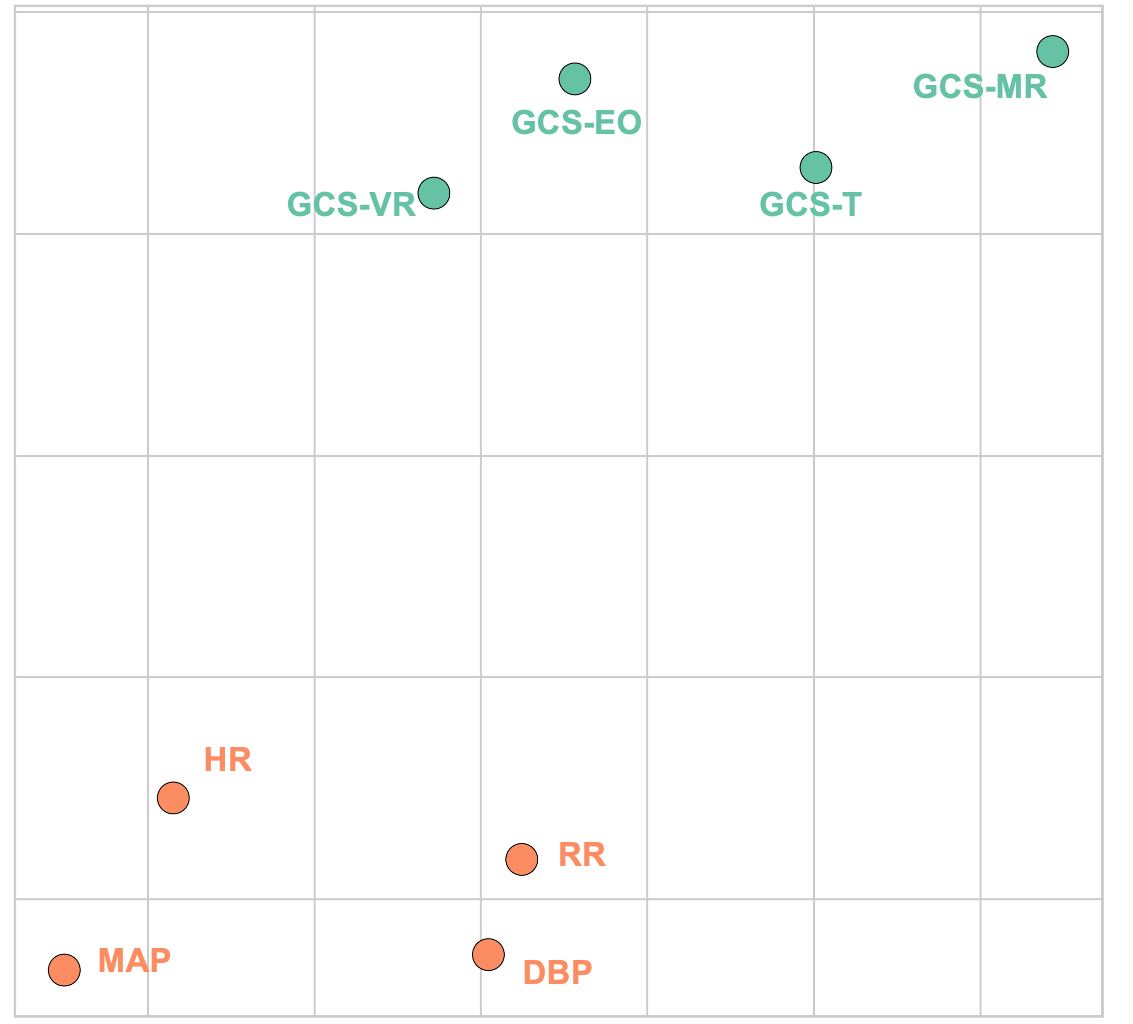}
        \caption{MIMIC-III Dataset}
        \label{fig:sub3}
    \end{subfigure}
    \caption{T-SNE visualization of partial variable representations on three datasets.}
    \label{fig:three_subfigs}
\end{figure}

Since DBGL builds a patient–variable bipartite graph, the learned variable nodes are expected to capture meaningful feature representations. To examine this, we visualize the variable embeddings using T-SNE \cite{maaten2008visualizing}. Following \cite{luo2024knowledge}, we group variables with similar temporal patterns and project their embeddings into a 2D space. The results, shown in Figure~\ref{fig:three_subfigs}, reveal clear clustering that aligns with clinical groupings, suggesting that DBGL effectively encodes variable-specific information. As the PhysioNet dataset is a subset of P12 and shares the same variables, we report the visualization only on P12.

Here, the temporal patterns include sampling rate, sampling time, observation span, sequence length, and trends. Variables sharing similar temporal patterns are grouped and visualized with the same color. We observe that, after training, embeddings of variables within the same group are tightly clustered, indicating that DBGL effectively captures the temporal dynamics of irregular clinical variables. This ability to model variable-specific temporal patterns contributes to more informative representations and improved classification performance.

\section{Experiments on PAM datasets}

To further evaluate the generalization of our proposed DBGL framework beyond binary clinical tasks, we conducted experiments on the PAM dataset \cite{reiss2012introducing}, which contains 18 physical activities across 9 subjects, with 17 variables (IMU + heart rate) sampled over 600 time steps, with 60\% missing ratio. 

We applied DBGL for classification. Unlike clinical datasets, PAM variables are dense and regularly sampled, with minimal variable-specific decay irregularity. Applying the decay-aware module here would introduce noise and degrade performance. 

\begin{table*}[t]
\centering
\caption{Performance comparison of different models on the dataset. Values are reported as mean $\pm$ std (\%).}
\begin{tabular}{lcccc}
\toprule
Model & Accuracy (\%) & Precision (\%) & Recall (\%) & F1 score (\%) \\
\midrule
Transformer  & 83.5 $\pm$ 1.5 & 84.8 $\pm$ 1.5 & 86.0 $\pm$ 1.2 & 85.0 $\pm$ 1.3 \\
Trans-mean   & 83.7 $\pm$ 2.3 & 84.9 $\pm$ 2.6 & 86.4 $\pm$ 2.1 & 86.4 $\pm$ 2.1 \\
GRU-D        & 83.3 $\pm$ 1.6 & 84.6 $\pm$ 1.2 & 84.6 $\pm$ 1.2 & 84.8 $\pm$ 1.2 \\
SeFT         & 67.1 $\pm$ 2.2 & 70.0 $\pm$ 2.4 & 68.2 $\pm$ 1.5 & 68.5 $\pm$ 1.8 \\
mTAND        & 74.6 $\pm$ 4.3 & 74.3 $\pm$ 4.0 & 79.5 $\pm$ 2.8 & 76.8 $\pm$ 3.4 \\
IP-Net       & 74.3 $\pm$ 3.8 & 75.6 $\pm$ 2.1 & 77.9 $\pm$ 2.2 & 76.6 $\pm$ 2.8 \\
DGM$^2$-O   & 82.4 $\pm$ 2.3 & 85.2 $\pm$ 1.2 & 83.9 $\pm$ 2.3 & 84.3 $\pm$ 1.8 \\
MTGNN        & 83.4 $\pm$ 1.9 & 85.2 $\pm$ 1.7 & 86.1 $\pm$ 1.9 & 85.9 $\pm$ 2.4 \\
RainDrop     & 88.5 $\pm$ 1.5 & 89.9 $\pm$ 1.5 & 89.9 $\pm$ 0.6 & 89.8 $\pm$ 1.0 \\
ViTST        & 95.8 $\pm$ 1.3 & 96.2 $\pm$ 1.3 & 96.2 $\pm$ 1.3 & 96.5 $\pm$ 1.2 \\
TimeCHEAT    & 96.5 $\pm$ 0.6 & 97.1 $\pm$ 0.5 & 96.9 $\pm$ 0.6 & 97.0 $\pm$ 0.5 \\
\hline
\rowcolor[gray]{0.95} DBGL         & 90.4 $\pm$ 0.9 & 92.5 $\pm$ 0.7 & 92.4 $\pm$ 0.9 & 92.4 $\pm$ 0.8 \\
\bottomrule
\end{tabular}
\label{tab:model_performance}
\end{table*}

DBGL achieves strong performance (92.4 F1), surpassing most irregular-time baselines. Its decay-aware module provides limited benefit here due to the lack of variable-wise decay heterogeneity, whereas ViTST and TimeCHEAT exploit strong generic sequence modeling (patching, global Transformers) suitable for dense continuous signals. 

Our decay-rate analysis confirms this: almost all variables exhibit extremely small decay values ($\lambda \approx 0.006\sim0.01$ ), indicating negligible heterogeneity. This aligns with the characteristics of PAM as dense motion sensor data without irregular sampling.

\begin{table}[h]
\centering
\caption{Variable-specific $\lambda$ values}
\begin{tabular}{l|c|l|c|l|c}
\toprule
Variable & $\lambda$ & Variable & $\lambda$ & Variable & $\lambda$ \\
\midrule
v0  & 1.0451 & v1  & 0.00887 & v2  & 0.00662 \\
v3  & 0.05265 & v4  & 0.00910 & v5  & 0.00660 \\
v6  & 0.04641 & v7  & 0.04618 & v8  & 0.00823 \\
v9  & 0.00720 & v10 & 0.00836 & v11 & 0.02725 \\
v12 & 0.01110 & v13 & 0.00728 & v14 & 0.00806 \\
v15 & 0.00756 & v16 & 0.00714 & —   & —       \\
\bottomrule
\end{tabular}
\label{tab:variable_lambda}
\end{table}

These results indicate that DBGL can generalize to diverse time series, even when the decay heterogeneity assumption does not hold.

\section{The usage of LLM}
We employed large language models (LLMs) in a limited capacity, strictly for writing assistance tasks such as grammar correction, style refinement, and table formatting. The LLM was also used solely to generate code comments and explanatory notes in the supplementary implementation, and no core experimental code was produced by the LLM. All main model implementations, data processing pipelines, and training routines were entirely written and verified by the authors. All suggested edits were carefully reviewed and selectively incorporated by the authors. The scientific content, ideas, analyses, and conclusions presented in this paper are entirely our own. The authors take full responsibility for the paper’s content, including any remaining errors or inaccuracies.

\end{document}